\documentclass[fullpaper,final,3p,times,twocolumn]{elsarticle}

\usepackage{lineno,hyperref}
\usepackage{fancyhdr, ifpdf}
\usepackage{amssymb}
\usepackage{amsmath}
\usepackage{mathrsfs}
\usepackage{color}
\usepackage{enumerate}
\usepackage{graphics}
\usepackage{epsfig}
\usepackage{psfrag}
\usepackage{amsfonts}
\usepackage{mathrsfs}
\usepackage{graphicx}
\usepackage{verbatim}
\usepackage{color}
\usepackage{booktabs}
\usepackage{epstopdf}
\usepackage[tight,footnotesize]{subfigure}
\usepackage{algorithm}
\usepackage{algorithmicx}
\usepackage{algpseudocode}
\usepackage{subfigure}
\usepackage{multicol}
\usepackage{multirow}
\usepackage{bbding}
\usepackage{caption2}

\modulolinenumbers[5]
\biboptions{sort&compress}

\begin{document}
\begin{frontmatter}

\title{Cluster-wise Unsupervised Hashing for Cross-Modal Similarity Search}

\author[mymainaddress]{Lu Wang}


\author[mymainaddress]{Jie Yang\corref{mycorrespondingauthor}}
\cortext[mycorrespondingauthor]{Corresponding author}
\ead{luwang\_16@sjtu.edu.cn, jieyang@sjtu.edu.cn}

\address[mymainaddress]{Institute of Image Processing and Pattern Recognition, Shanghai Jiao Tong University, Shanghai, China}

\begin{abstract}
Large-scale cross-modal hashing similarity retrieval has attracted more and more attention in modern search applications such as search engines and autopilot, showing great superiority in computation and storage. However, current unsupervised cross-modal
hashing methods still have some limitations: (1)many methods relax the discrete constraints to solve the optimization objective which may significantly degrade the retrieval performance;(2)most existing hashing model project heterogenous data into a common latent space, which may always lose sight of diversity in heterogenous data;(3)transforming real-valued data point to binary codes always results in abundant loss of information, producing the suboptimal continuous latent space. To overcome above problems, in this paper, a novel Cluster-wise Unsupervised Hashing (CUH) method is proposed. Specifically, CUH jointly performs the multi-view clustering that projects the original data points from different modalities into its own low-dimensional latent semantic space and finds the cluster centroid points and the common clustering indicators in its own low-dimensional space, and learns the compact hash codes and the corresponding linear hash functions. An discrete optimization framework is developed to learn the unified binary codes across modalities under the guidance cluster-wise code-prototypes. The reasonableness and effectiveness of CUH is well demonstrated by comprehensive experiments on diverse benchmark datasets.
\end{abstract}


\end{frontmatter}

\section{Introduction}
Due to the explosive growth of big data with multiple modalities in the form of images, text, and videos on social networks, efficient data analysis has gotten an immediate attention to purify the semantic correlations across different heterogenous modalities. In other word, when we have relevant data in different modalities endowing the semantic correlation structures, it always is desirable to perform cross-modal search, which retrieves the semantically -similar items across the heterogeneous modalities in response to a query. Taking Wikipedia as an example, we can retrieval images of a relevant query tag, or tags of a relevant query image. Nevertheless, as a result of large-scale databases, heterogeneity, diversity and huge semantic gap, it still remains a great challenge for effective and efficient cross-modal retrieval.

Under the circumstances that the searchable database has large volume or that the similarity measure calculation between query item and database items is expensive, hashing based methods gets great popularity for its low storage cost, fast searching speed and impressive retrieval performance. Moreover, a hash method will search approximate nearest neighbor (ANN) within the reference database for a query item in many tasks such as machine learning \cite{proceeding21, proceeding22}, data mining \cite{jour7, jour8} and computer vision \cite{jour9, jour10}, which could balance retrieval efficiency against retrieval accuracy. The basic principle for hashing is to transform each high-dimensional data point into compact binary code, making close binary codes for the relevant data samples in different modalities.

In recent time, various kinds of attempts have been investigated for cross-modal hashing, which encodes the correlation structures between different heterogeneous modalities when learning hash function and indexing cross-modal data points in the Hamming space \cite{proceeding6, proceeding7, proceeding8}, \cite{proceeding11}, \cite{proceeding1, proceeding2, proceeding3, proceeding4, proceeding5}. These existing cross-modal hashing methods always can be induced to a two-step scheme: first, projected multiple heterogenous data modalities into a continuous common latent space by optimizing inter-modal coherence, and second, quantize the continuous projections into compact binary codes by sign function. While demonstrating successful performance, there are some limitation in the two-step scheme: first, transformation from real-valued data to discrete binary codes always results in abundant loss of information, producing the suboptimal continuous common latent space and the suboptimal compact binary codes \cite{proceeding6, proceeding7}; second, solving the optimization objective by relaxing the discrete constraints which may significantly degrade the retrieval performance with great quantization error \cite{proceeding1, proceeding2}; third, projecting heterogenous data into a common latent space can always lose sight of diversity, which could help to learn better binary codes for cross-modal search to some degree. Hence, how to learn compact binary codes with excellent performance is still a great challenge work. Besides, generally speaking, we can roughly classify existing cross-modal hashing methods into unsupervised ones \cite{proceeding6, proceeding7, proceeding8}, \cite{proceeding11} and supervised ones \cite{proceeding1, proceeding2, proceeding3, proceeding4, proceeding5}. The details are represented in the Section 2.
\begin {figure*}[t]
\centering
\includegraphics[width=0.9\textwidth]{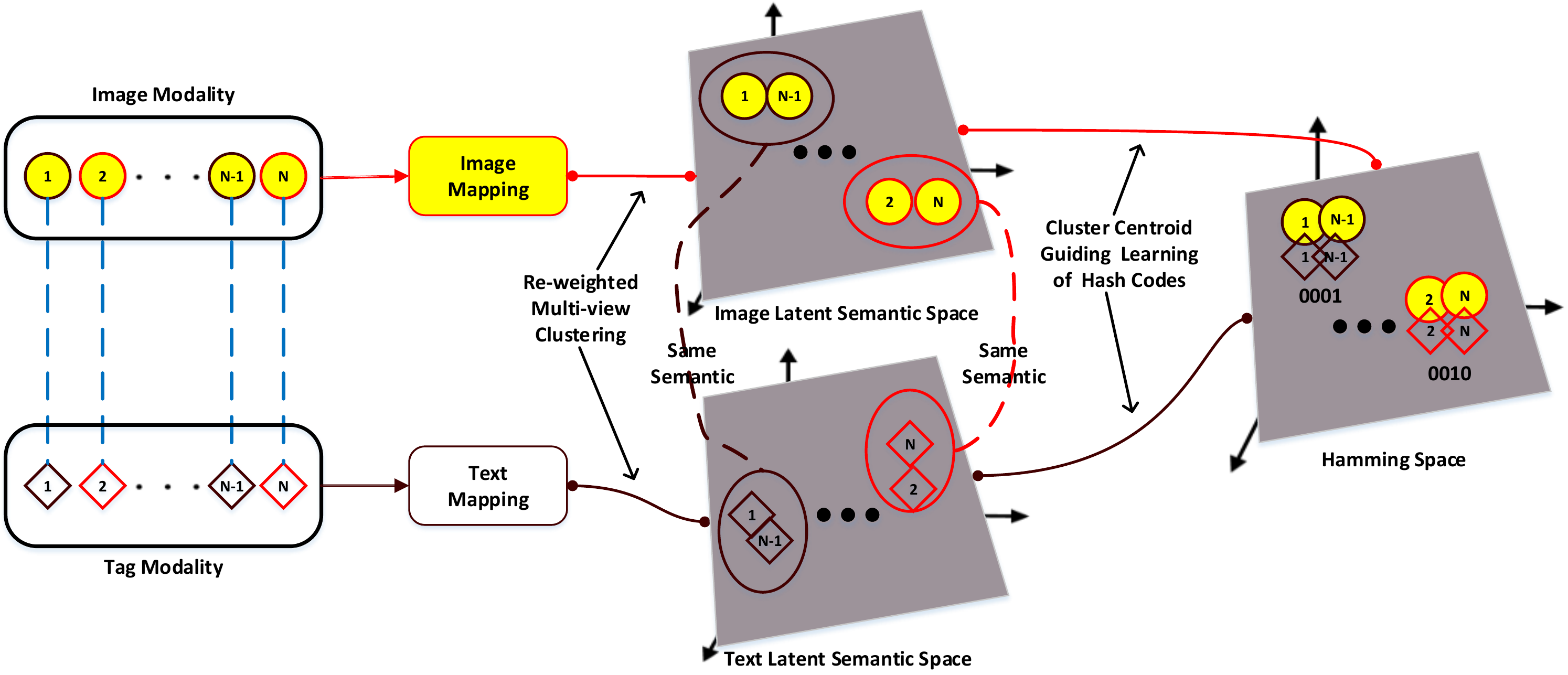}
\caption{The flowchart of CUH.}
\label{figure100}
\end {figure*}

In this paper, we propose Cluster-wise Unsupervised Hashing (CUH), a novel hash model performing effective and efficient cross-modal retrieval. Technically, CUH jointly performs the multi-view clustering that projects the original data points from different modalities into its own low-dimensional latent semantic space and finds the cluster centroid points and the common clustering indicators in its own low-dimensional space, and learns the compact hash codes and the corresponding linear hash functions. The flowcharts of CUH are shown in Fig. \ref{figure100}. To construct a seamless learning framework, we are inspired  by the work of class-wise supervised hashing \cite{proceeding16} and the work of re-weighted discriminatively embedded K-means for multi-view clustering \cite{jour6}, and create a co-training framework for learning to hash in the unsupervised case, in which we simultaneously realize the multi-view clustering, the learning of hash codes and the learning of hash functions. These above steps are jointly optimised by a unified learning problem, which could keep both inter-modal semantic coherence and intra-modal similarity when minimizing both the multi-view least-absolute clustering residual and the quantization error. The CUH model can generates one extremely compact unified hash code to all observed modalities of any instance for efficient cross-modal search and could scale linearly to the data point size. The reasonableness and effectiveness of CUH is well demonstrated by comprehensive experiments on diverse benchmark datasets.

We summarize the contributions of this paper as follows.
\begin{enumerate}[1)]
\item We propose a cluster-wise unsupervised hashing method, which constructs a co-training framework for learning to hash. In the framework, we simultaneously realize the multi-view clustering and the learning of hash codes.
\item We propose a alternately optimization scheme for solving our model. Besides, we develop a discrete optimization method to jointly learn binary codes and the corresponding hash functions for each modality which can improve the performance.
\end{enumerate}

The remainder of this paper is structured as follows. In Section 2, we briefly overview the related works of cross-modal hashing methods. Section 3 elaborates our proposed cluster-wise unsupervised hashing method, along with an efficient discrete optimization algorithm to tackle this problem. In Section 4, we report the experimental results and extensive evaluations on popular benchmark datasets. Finally, we draw a conclusion in Section 5.
\section{Related work}
As mentioned above, there are two categories cross-modal hashing methods, i.e. unsupervised and supervised ones. The former ones
maximize intra-modality and inter-modality relevance of the features of training data for learning hash functions. Meanwhile, the latter ones can better learn the hash functions and acquire superior performance by further utilizing the available supervised information. Actually, for supervised methods, they usually require label information of the entire data, which is difficult when the database is large-scale. Recently, deep learning based cross-modal hashing methods have attracted increasing attention for their significant performance improvements, where an end-to-end deep learning architecture can give binary codes for different modalities, capturing the intrinsic cross-modal relevance \cite{proceeding13, proceeding14, proceeding15}.

IMH \cite{proceeding8}, SMMH \cite{proceeding7}, CMFH \cite{proceeding11}, LSSH \cite{proceeding6} are unsupervised cross-modal hashing methods. Song et al. proposed inter-media hashing (IMH), which maximizes the intra-modality and inter-modality consistencies for learning binary codes \cite{proceeding8}. Zhen et al. proposed spectral multi-modal hashing (SMMH), which is an extension of spectral analysis of the correlation matrix to obtain binary hash codes \cite{proceeding7}. Ding et al. proposed collective matrix factorization hashing (CMFH) that performs collective matrix factorization to learn unified hash codes \cite{proceeding11}. Zhou et al. proposed latent semantic sparse hashing (LSSH), which respectively, utilizes sparse coding for images and matrix factorization for texts to learn their latent semantic features to generate unified hash codes \cite{proceeding6}.

Differently, CMSSH \cite{proceeding5}, CVH \cite{proceeding3}, CRH \cite{proceeding4}, DCDH \cite{proceeding2}, SCM \cite{proceeding1} are supervised cross-modal hashing methods. Bronatein et al. proposed cross-modality similarity-sensitive hashing (CMSSH) that models the binary classification problems for the projections from features in each modality to hash codes, and utilizes boosting methods to efficiently learn them \cite{proceeding5}. Kumar et al. proposed a cross-view hashing (CVH), which is an extension of the single-modal spectral hashing \cite{proceeding3}. Zhen et al. proposed co-regularized hashing (CRH) to learn hash function for multi-modal data in a boosted co-regularization framework \cite{proceeding4}. Yu et al. proposed discriminative coupled dictionary hashing (DCDH) \cite{proceeding2}, which learns a coupled dictionary for each modality and unified hash functions. Zhang et al. proposed semantic correlation maximization (SCM) to take semantic labels into consideration for the hash learning procedure in large-scale datasets \cite{proceeding1}.

In recent years, deep learning methods have acquired great performance improvements on various tasks \cite{proceeding13, proceeding14, proceeding15}. Inspiring from the advancement of deep learning, many cross-modal hashing methods have developed significant frameworks with deep neural networks, which can bridge the heterogeneous modalities more effectively by overcoming the insufficient character of the hand-crafted features. While these deep models can lead to outstanding performance, there also are some problem such as computational complexity and exhaustive search of learning parameters. Besides, another limitation is that these approaches cannot well reduce the gap between the Hamming distance and the metric distance on real-valued high-level data representations.

After surveying the existing cross-modal hashing methods, we can clear that well preserving the semantic relevance between instances is the key for reducing the quality loss when retrieving neighbors and achieving better performance. Therefore, it is still desirable to develop a flexible cross-modal retrieval algorithm. Differently, in this paper, the proposed CUH further incorporates the correlations between pairwise Hamming distances to force the to-be-learnt hash codes to better preserve the semantic relevance. As will be demonstrated by our experiments, CUH is reasonable and emerges superior performance.
\section{Proposed Algorithm}
In this section,we will present the detail of the CUH algorithm.
\subsection{Notations and Problem Statements}
Now, we describe in details the cross-modal retrieval system. Let the two modalities be denoted as $X_1=\left[x_1^1,x_1^2,\ldots,x_1^N \right]\in \Re^{d_1\times N}$ and $X_2=\left[x_2^1,x_2^2,\ldots, x_2^N \right] \in \Re^{d_2\times N}$, with $N$ being the number of items in either modality and $d_1, d_2$ being the dimensionality of the data (in general $d_1\ne d_2$) respectively. Without loss of generality, we assume the input instances in $X_1$ and $X_2$ are both zero centered, i.e., $\sum_{i=1}^{N} (X_v)_i=0, v=1, 2$.

Given such data, the goal of CUH is to learn the unified binary codes matrix $B=\left\{b_i\right\}_{i=1}^N \in \left\{+1, -1\right\}^{r\times N}$ for training instances in both $X_1$ and $X_2$. Besides, $b_i \in \left\{+1, -1\right\}^{r}$ is the unified $r$-bits binary codes vector for both instance $x_1^i$ and $x_2^i$. The modal-specific hash functions aims to map each input instances from corresponding modality to a binary code of $r$ bits through learning $r$ hash functions as follows,
\begin{equation}\label{hashfunction}
\begin{split}
H_1(x_1^i)=sgn(W_1^Tx_1^i), \\
H_2(x_2^j)=sgn(W_2^Tx_2^j),
\end{split}
\end{equation}
where $H_1(\cdot)$ and $H_2(\cdot)$ are modal-specific hash functions for image and text modalities, respectively, $x_1^i$ is the $i$-th input instance from the image modality, $x_2^j$ is the $j$-th input instance from the text modality. Here, $W_1\in \Re^{d_1\times r}$ and $W_2\in \Re^{d_2\times r}$ are the linear projection matrices that map the original feature of $x_1^i$ and $x_2^j$ to low-dimensional latent spaces, respectively. The sign function $sgn(\cdot)$ outputs $+1$ for positive number and $-1$ otherwise.
\subsection{Cluster-wise Unsupervised Hashing}
The main framework of CUH is to jointly find the cluster centroid points and the common clustering indicators in its own low-dimensional semantic space and learn the unified hash codes under the guidance of the cluster centroid points, where the cluster centroid points are the cluster-wise code-prototypes to improve the performance of the binary codes.

To realize this mission, we are inspired by the work of re-weighted discriminatively embedded K-means for multi-view (image and text) clustering \cite{jour6} and induce a robust re-weighted discriminatively embedded K-means for maximizing the inter-modality consistencies to get the cluster centroid points and the common clustering indicators in its own low-dimensional semantic space. In the process for learning of hash codes, we utilize the cluster centroid points as the cluster-wise code-prototypes to guide the learning of the corresponding hash codes of original data for more precise compact codes for semantic information retrieval. We describe how to construct the CUH method under above idea.
\subsubsection{Multi-view Clustering}
In order to deal with multi-view and high-dimensional data, re-weighted discriminatively embedded K-means proposes an objective function as follows,
\begin{eqnarray}\label{rdekm}
\min_{W_k, F_k, G} & & \sum_{k=1}^2 \left \| W_k^TX_k-F_kG^T \right \|_F \nonumber \\
\mathrm{s.t.} & & W_k^TW_k=I_{m_k}, k=1,2,  \\
& & G\in Ind, \nonumber
\end{eqnarray}
where $W_k\in \Re^{d_k\times m_k}$ represents the projection matrix which reduces the dimensionality from $d_k$ to $m_k$ for each view, $F_k\in \Re^{m_k\times C}$ is the cluster centroid matrix and each column of $G$ denotes the clustering indicator vector for each sample where $G_{ic}=1 (i=1, \ldots, N; c=1, \ldots, C)$ if the $i$-th sample belongs to the $c$-th class and $G_{ic}=0$ otherwise. Thus, $G\in Ind$ can be defined, which denotes a set of matrices with above restrictions. For adaptively learning the weights in a re-weighted manner, the objective function can be defined as:
\begin{eqnarray}\label{rdekm2}
\min_{W_k, F_k, G, \alpha_k} & & \sum_{k=1}^2 \alpha_k \left \| W_k^TX_k-F_kG^T \right \|_F^2 \nonumber \\
\mathrm{s.t.} & & W_k^TW_k=I_{m_k}, k=1,2,  \\
& & G\in Ind, \nonumber
\end{eqnarray}
where $\alpha_k=(2\left \| W_k^TX_k-F_kG^T \right \|_F)^{-1}$ is the weight for the $k$-th view and can be calculated by current $W_k$, $F_k$ and $G$.
\subsubsection{Learning of Hash Codes under Cluster-wise Code-prototypes}
We can get hash codes from above multi-view clustering in a co-training framework. In the process for learning of hash codes, the dimension reduced data is used as the approximation for the corresponding hash codes of original data. Besides, the cluster centroid points are the cluster-wise code-prototypes. These cluster-wise code-prototypes can guide the learning of the corresponding hash codes of original data to improve the performance of the binary codes. For this goal, we come up with the following objective function,
\begin{eqnarray}\label{LHC}
\min_{B} & & \sum_{k=1}^2 \left \| B-W_k^TX_k \right \|_F^2 - \beta tr(B^TF_kG^T)\nonumber \\
\mathrm{s.t.} & & B \in \left\{+1, -1\right\}^{r\times N},
\end{eqnarray}
where $\beta$ is the parameter to balance the reconstruction error and the similarity between cluster-wise code-prototypes and binary codes.
\subsubsection{Joint Optimization Framework}
To approach CUH, which jointly finds the cluster centroid points and the common clustering indicators in its own low-dimensional semantic space and learns the unified hash codes under the guidance of the cluster centroid points, we combine the aforementioned description. That can  bring about the objective function of CUH is written below:
\begin{eqnarray}\label{CUH}
\min_{W_k, F_k, G, \alpha_k, B} & & \sum_{k=1}^2 (\alpha_k \left \| W_k^TX_k-F_kG^T \right \|_F^2 \nonumber \\
& & + \lambda \left \| B-W_k^TX_k \right \|_F^2 - \beta tr(B^TF_kG^T)) \nonumber \\
\mathrm{s.t.} & & W_k^TW_k=I_{r}, k=1,2,  \\
& & B \in \left\{+1, -1\right\}^{r\times N}, \nonumber \\
& & G\in Ind, \nonumber
\end{eqnarray}
By minimizing (\ref{CUH}), the unified hash binary codes will be obtained directly.
\subsection{Optimization}
To find a feasible solution for the optimization problem (\ref{CUH}), in this section, we present an alternating optimization approach.
\begin{enumerate}[1)]

\item $W_k$ and $F_k$-step: fix $G$, $\alpha_k$ and $B$, update $W_k$ and $F_k$

By fixing $G$, $\alpha_k$ and $B$, the optimization problem (\ref{CUH}) becomes
\begin{eqnarray}\label{CUH2}
\min_{W_k, F_k} & & \sum_{k=1}^2 (\alpha_k \left \| W_k^TX_k-F_kG^T \right \|_F^2 \nonumber \\
& & + \lambda \left \| B-W_k^TX_k \right \|_F^2 - \beta tr(B^TF_kG^T)) \nonumber \\
\mathrm{s.t.} & & W_k^TW_k=I_{r}, k=1,2.
\end{eqnarray}

Calculating $W_k$ and $F_k$ is a supervised learning stage when we fix $G$, $\alpha_k$ and $B$. Firstly, we rewrite Eq.(\ref{CUH2}) as following Eq.(\ref{CUH3}) which is very to implement and can be readily used to solve a general trace minimization problem:
\begin{eqnarray}\label{CUH3}
\min_{W_k} & & \sum_{k=1}^2 (tr(W_k^TM_kW_k)-2tr(W_k^TN_k)) \nonumber \\
\mathrm{s.t.} & & W_k^TW_k=I_{r}, k=1,2,
\end{eqnarray}
where $M_k=(\alpha_k+\lambda)X_kX_k^T-\alpha_kX_kG(G^TG)^{-1}G^TX_k^T$, $N_k=\lambda X_kB^T+\frac{\beta}{2} X_kG(G^TG)^{-1}G^TB^T$. Besides, we obtain $F_k=(\frac{\beta}{2\alpha_k} B +W_k^TX_k)G(G^TG)^{-1}$.

We can use the orthogonal constraint optimization procedure in \cite{jour4}, \cite{proceeding12}. Through introducing Lagrangian multipliers, we can rewrite the objective function for optimizing $W_k(k=1,2)$ as follows:
\begin{equation}\label{CUH4}
\begin{split}
L(W_k,\Lambda)=&tr(W_k^TM_kW_k)-2tr(W_k^TN_k)\\
&-tr(\Lambda (W_k^TW_k-I)),
\end{split}
\end{equation}
where $\Lambda$ consists of Lagrangian multipliers. Since $W_k^TW_k$ is symmetric, $\Lambda$ is symmetric as well. Setting the gradient of Eq.(\ref{CUH4}) with respect to $W_k$ to be zero, we can get
\begin{equation}\label{CUH5}
\frac{\partial L(W_k,\Lambda)}{\partial W_k}=2(M_kW_k-N_k-W_k\Lambda)=0.
\end{equation}
From Eq.(\ref{CUH5}), it is clear that we can get $\Lambda=W_k^TM_kW_k-W_k^TN_k$. So $\Lambda=W_k^TM_kW_k-W_k^TN_k=W_k^TM_kW_k-N_k^TW_k$ and $\frac{\partial L(W_k,\Lambda)}{\partial W_k}=2(M_kW_k-N_k-W_kW_k^TM_kW_k+W_kN_k^TW_k)$. Based on the orthogonal constraint optimization procedure in \cite{jour4}, we can define a skew-symmetric matrix $A=2(M_kW_kW_k^T-N_kW_k^T-W_k^TM_kW_k+N_k^TW_k)$. Then, we will update $W_k$ by Crank-Nicolsonlike scheme \cite{jour5}
\begin{equation}\label{CUH6}
W_k^{(t+1)}=W_k^{(t)}-\frac{\tau}{2} A(W_k^{(t+1)}+W_k^{(t)}),
\end{equation}
where $\tau$ is the step size. By solving (\ref{CUH6}), we can obtain
\begin{equation}\label{CUH7}
\begin{split}
W_k^{(t+1)}=QW_k^{(t)}, \\
Q=(I+\frac{\tau}{2} A)^{-1} (I-\frac{\tau}{2} A).
\end{split}
\end{equation}
Hereafter, we iteratively update $W_k$ several times based on Eq.(\ref{CUH7}) with Barzilai-Borwein (BB) method \cite{jour4}. In addition, please note that when iteratively optimizing $W_k$, the initial $W_k$ is set to be the one optimized in the last round between B and $W_k$. For the first round,$W_k$ is randomly initialized.
\item $G$-step: fix $W_k$, $F_k$, $\alpha_k$ and $B$, update $G$
By fixing $W_k$, $F_k$, $\alpha_k$ and $B$, the optimization problem (\ref{CUH}) becomes
\begin{eqnarray}\label{CUH10}
\min_{G} & & \sum_{k=1}^2 (\alpha_k \left \| W_k^TX_k-F_kG^T \right \|_F^2 \nonumber \\
& & - \beta tr(B^TF_kG^T)) \nonumber \\
\mathrm{s.t.} & & G\in Ind.
\end{eqnarray}

Obtaining the clustering indicator matrix $G$ via a weighted multi-view K-Means clustering is an unsupervised learning stage. We search the optimal solution of $G$ among multiple low-dimensional discriminative subspaces. By separating $X_k$ and $G$ into independent vectors respectively, Eq.(\ref{CUH10}) can be replaced by the following problem:
\begin{eqnarray}\label{CUH11}
\min_{G} & & \sum_{k=1}^2 (\alpha_k \left \| W_k^TX_k-F_kG^T \right \|_F^2 \nonumber \\
& & - \beta tr(B^TF_kG^T)) \nonumber \\
=\min_{G} & & \sum_{i=1}^N \sum_{k=1}^2 (\alpha_k \left \| W_k^Tx_k^i-F_kg_i^T \right \|_F^2 \nonumber \\
& & - \beta b_i^TF_kg_i^T) \nonumber \\
=\min_{G} & & \sum_{i=1}^N \sum_{k=1}^2 \alpha_k \left \| W_k^Tx_k^i+\frac{\beta}{2\alpha_k} b_i-F_kg_i^T \right \|_F^2 \nonumber \\
\mathrm{s.t.} & & G\in Ind, g_i \in G, \nonumber \\
& & g_{ic} \in \left\{0, 1\right\}, \sum_{c=1}^C g_{ic}=1,
\end{eqnarray}
where $g_i$ is the $i$-th row of $G$ which denotes the clustering indicator vector for the $i$-th sample. Moreover, $g_{ic}$ denotes the $c$-th element of $g_i$, and there are $C$ candidates to be $g_i$ and each of them is the $c$-th row of identity matrix $I_C$:
\begin{equation}
I_C=\left[e_1, e_2, \ldots , e_C \right], g_i \in \left\{e_1^T, e_2^T, \ldots , e_C^T \right\}. \nonumber
\end{equation}

The one among C candidates making the objective function reach the minimum value is the solution of Eq.(\ref{CUH11}). We solve Eq.(\ref{CUH11}) by separating data matrix $X_k$ along the data points direction and assigning $C$ different $e_c^T$ to the row vector $g_i$ one by one independently. Thus, we can tackle the following problem for the $i$-th sample:
\begin{equation}\label{CUH12}
c^*=arg \min_{c} \sum_{k=1}^2 \alpha_k \left \| W_k^Tx_k^i+\frac{\beta}{2\alpha_k} b_i-F_ke_c \right \|_F^2,
\end{equation}
where $c^*$ means that the $c$-th element of $g_i$ is $1$ and others are $0$. There are only $C$ kinds of candidate clustering indicator vector, so we can easily find out the solution of Eq.(\ref{CUH11}).
\item $\alpha_k$-step: fix $W_k$, $F_k$, $G$ and $B$, update $\alpha_k$
By fixing $W_k$, $F_k$, $G$ and $B$, Updating the non-negative weight $\alpha_k$ for each view assigns the more discriminative image feature with higher weight. The $W_k$ and $G$ in the $t$-th iteration are computed from the solution of the current iteration. With the current $W_k^{(t)}$ and $G^{(t)}$, we can derive the closed form solution for $\alpha_k^{(t+1)}$£º
\begin{equation}\label{CUH13}
\alpha_k^{(t+1)}=(2\left \| W_k^{(t)T}X_k-F_k^{(t)}G^{(t)T} \right \|_F)^{-1}.
\end{equation}
Note that $W_k^{(t)}$ and $G^{(t)}$ are independent of $\alpha_k^{(t)}$ and can be considered as the constants. We iteratively solve $\alpha_k^{(t+1)}$ based on current $W_k^{(t)}$ and $G^{(t)}$.
\item $B$-step: fix $W_k$, $F_k$, $G$ and $\alpha_k$, update $B$
By fixing $W_k$, $F_k$, $G$ and $\alpha_k$, the optimization problem (\ref{CUH}) becomes
\begin{eqnarray}\label{CUH14}
\min_{B}  Q(B)= & & \sum_{k=1}^2 (\lambda \left \| B-W_k^TX_k \right \|_F^2 \nonumber \\
& & - \beta tr(B^TF_kG^T)) \nonumber \\
\mathrm{s.t.} & & B \in \left\{+1, -1\right\}^{r\times N}.
\end{eqnarray}

To solve this optimization problem, we further rewrite above problem (\ref{CUH14}) as follows,
\begin{equation}
\begin{split}
Q(B)= \sum_{k=1}^2 (\lambda (\left \| B\right \|_F^2+ \left \| W_k^TX_k \right \|_F^2-2tr(B^TW_k^TX_k)) \\
- \beta tr(B^TF_kG^T)). \nonumber
\end{split}
\end{equation}
Since $\left \| B\right \|_F^2$ and $\left \| W_k^TX_k \right \|_F^2$ are both constant, we have
\begin{eqnarray}\label{CUH15}
Q(B) = & & \sum_{k=1}^2 (-2\lambda tr(B^TW_k^TX_k)\nonumber \\
& &  - \beta tr(B^TF_kG^T))+const \nonumber \\
= & & -2tr(V^TB)+const,
\end{eqnarray}
where $V=\sum_{k=1}^2 (\lambda W_k^TX_k+\frac{\beta}{2} F_kG^T)$. Minimizing $Q(B)$ is equivalent to maximizing $tr(V^TB)$. As $B \in \left\{+1, -1\right\}^{r\times N}$, the optimal solution for (\ref{CUH14}) can be obtained by setting
\begin{equation}\label{CUH16}
B=sgn(\sum_{k=1}^2 (\lambda W_k^TX_k+\frac{\beta}{2} F_kG^T)).
\end{equation}
\end{enumerate}

To sum up, by these four steps, we can alternatively update $W_k, F_k, G, \alpha_k$ and $B$ and iterate the procedure above until the objective function get a stable minimum value. The process of CUH can be outlined in Algorithm \ref{alg:CUH}.
\begin{algorithm}[h]
  \caption{Cluster-wise Unsupervised Hashing.}
  \label{alg:CUH}
  \begin{algorithmic}[1]
    \Require
      feature matrices $X_1$ and $X_2$, code length $r$, parameters $\lambda$ and $\beta$.
    \Ensure
      hash codes $B$, $W_1$ and $W_2$.
    \State Initialize $W_1, W_2$ by identity.
    \State Initialize $G \in Ind$ randomly.
    \State Initialize binary codes $B$ randomly, such that $+1$ and $-1$ are balanced in the codes.
    \State Initialize the weight $\alpha_1=0.5, \alpha_2=0.5$ for each modality.
    \Repeat
      \State Update $W_k$, $(t=1,2)$, by Eq.(\ref{CUH7}) and obtain $F_k$, $(t=1,2)$, by:
      \State $F_k=(\frac{\beta}{2\alpha_k} B +W_k^TX_k)G(G^TG)^{-1}$.
      \State Update $G$ by Eq.(\ref{CUH12}).
      \State Update binary codes $B$ by Eq.(\ref{CUH16}).
      \State Update the weight  $\alpha_t(t=1,2)$ by Eq.(\ref{CUH13}).
    \Until Objective function of Eqn.(\ref{CUH}) converges.
  \end{algorithmic}
\end{algorithm}
\subsection{Generating Hash Codes for Queries}
Given a new query instance $x_k^q (k=1, 2)$, generating its binary codes $b^q$ depends on its modality. When $x_k^q (k=1, 2)$ contains data of only one modality, it is straightforward to predict its unified binary codes via the modality-specific hash function. When $x_k^q (k=1, 2)$ contains data of both two modalities, its unified binary codes are determined by merging the predicted binary codes from different modalities. Thus, the binary codes generation scheme for $x_k^q (k=1, 2)$ includes the following two situations:

Only one modality. In this case, we have $x_1^q$ or $x_2^q$. For $x_k^q (k=1, 2)$, we directly compute its binary codes $b^q$ as $b^q=sgn(W_k^Tx_k^q)$.

Two modalities. In this case, we both have $x_1^q$ and $x_2^q$. For CUH, we add up the results computed by the hash functions of two modalities and generate $b^q$ as $b^q=sgn(W_1^Tx_1^q+W_2^Tx_2^q)$.
\subsection{Complexity Analysis}
We discuss the computational complexity of the proposed CUH. In the training phase, the time consuming of each iteration including updating the projection matrices $W_1$ and $W_2$, the cluster centroid matrix $F_1$ and $F_2$, the clustering indicator matrix $G$, the binary codes $B$ and the weight  $\alpha_1$ and $\alpha_1$. Typically, solving Eq.(\ref{CUH7}), calculating $F_1$ and $F_2$, solving Eq.(\ref{CUH12}), solving Eq.(\ref{CUH16}) and solving Eq.(\ref{CUH13}) require $O (d_k^2r+d_k^3)$, $O (rN+rCN+C^2N+C^3+rC^2)$, $O (rd_kN+rCN+rN+CN)$, $O (rd_kN+rCN+rN)$ and $O (rd_kN+rCN+rN)$. Therefore, the time complexity of each iteration is $O (f_1d_k^2+f_2N+f_3C^2)$, where $f_1=\max(r, d_k)$, $f_2=\max(r, rC, C^2, rd_k,r, C)$ and $f_3=\max(r, C^2)$. The time complexity of all iterations is $O ((f_1d_k^2+f_2N+f_3C^2)T)$, where $T$ is the number of iterations. It can be observed that the training time is linear to the training set size $N$. Besides, in the experiments part, we will show that CUH usually only needs few iterations (T is very small) to achieve the best modal parameters. Once the training stage is done, the time and space complexities for generating binary codes for a new query are both $O (d_kr)$ in the query stage, which is extremely efficient. In general, CUH is scalable for large-scale data sets with most existing cross-modal hashing methods and efficient for encoding new query.
\section{Experiments and evaluations}
In this section, we execute comprehensive retrieval performance evaluation of CUH on three multimodal benchmark data sets against several state-of-the-art unsupervised cross-modal hashing methods. We present the details of concrete content in data sets, evaluation criteria, comparison methods, and implementation details for the first time. Next, We investigate the experimental results and discussions in terms of fair comparisons. Finally, the convergence and parameter sensitivity of CUH are further reported.
\subsection{Data Sets}
\begin {figure} [t]
\centering
\subfigure{
  \includegraphics[width=.2\textwidth]{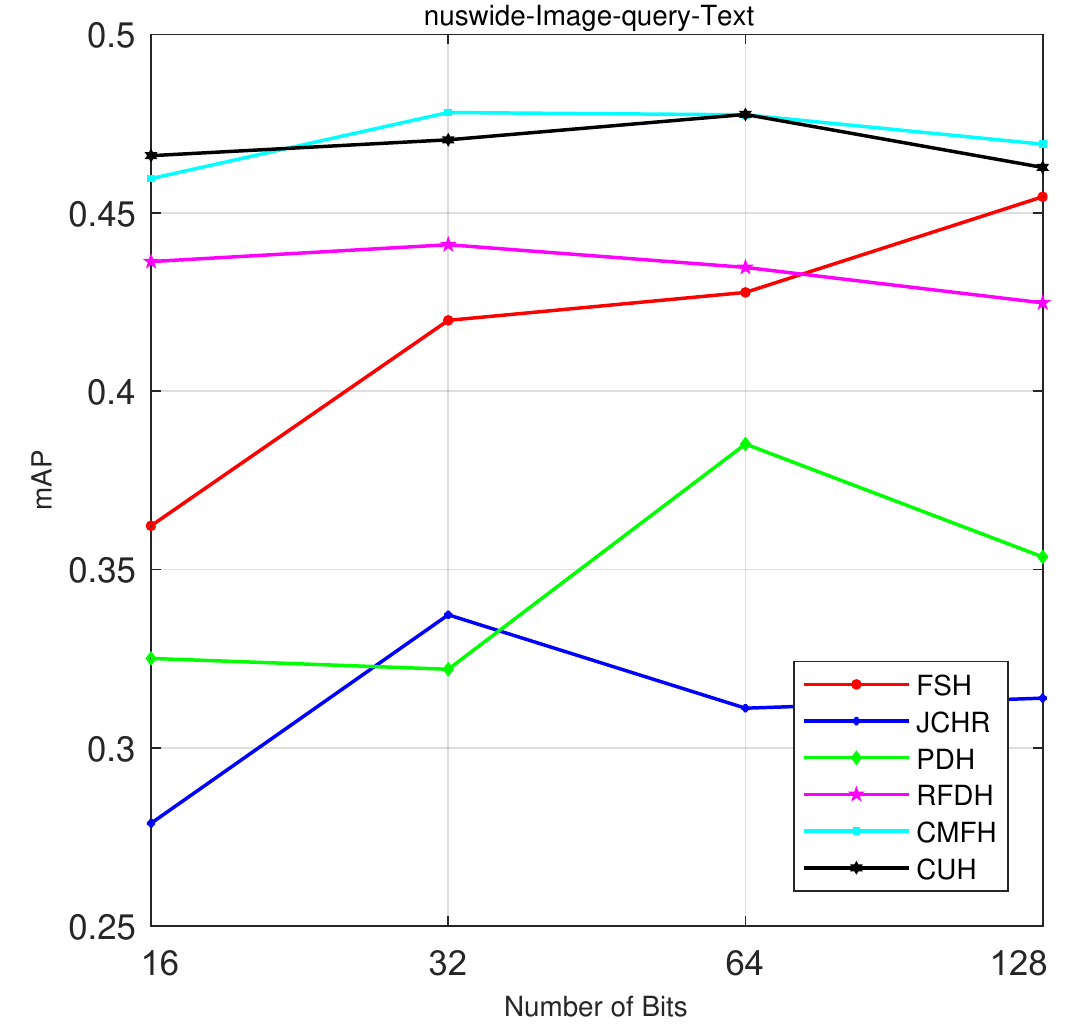}}
\hfill
\centering
\subfigure{
  \includegraphics[width=.2\textwidth]{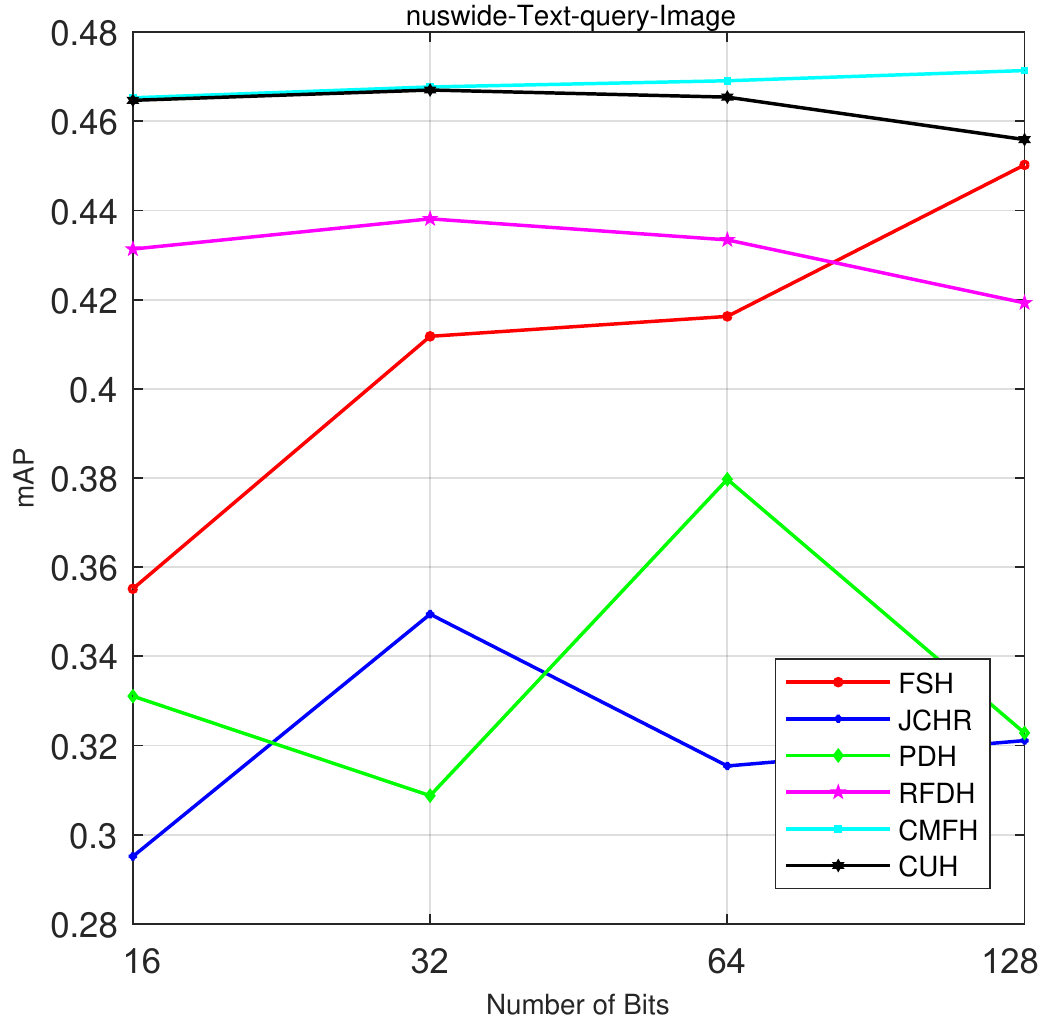}}
\caption{mAP values versus bits on Nuswide.}
\label{figure5}
\end {figure}
\begin {figure} [t]
\centering
\subfigure{
  \includegraphics[width=.2\textwidth]{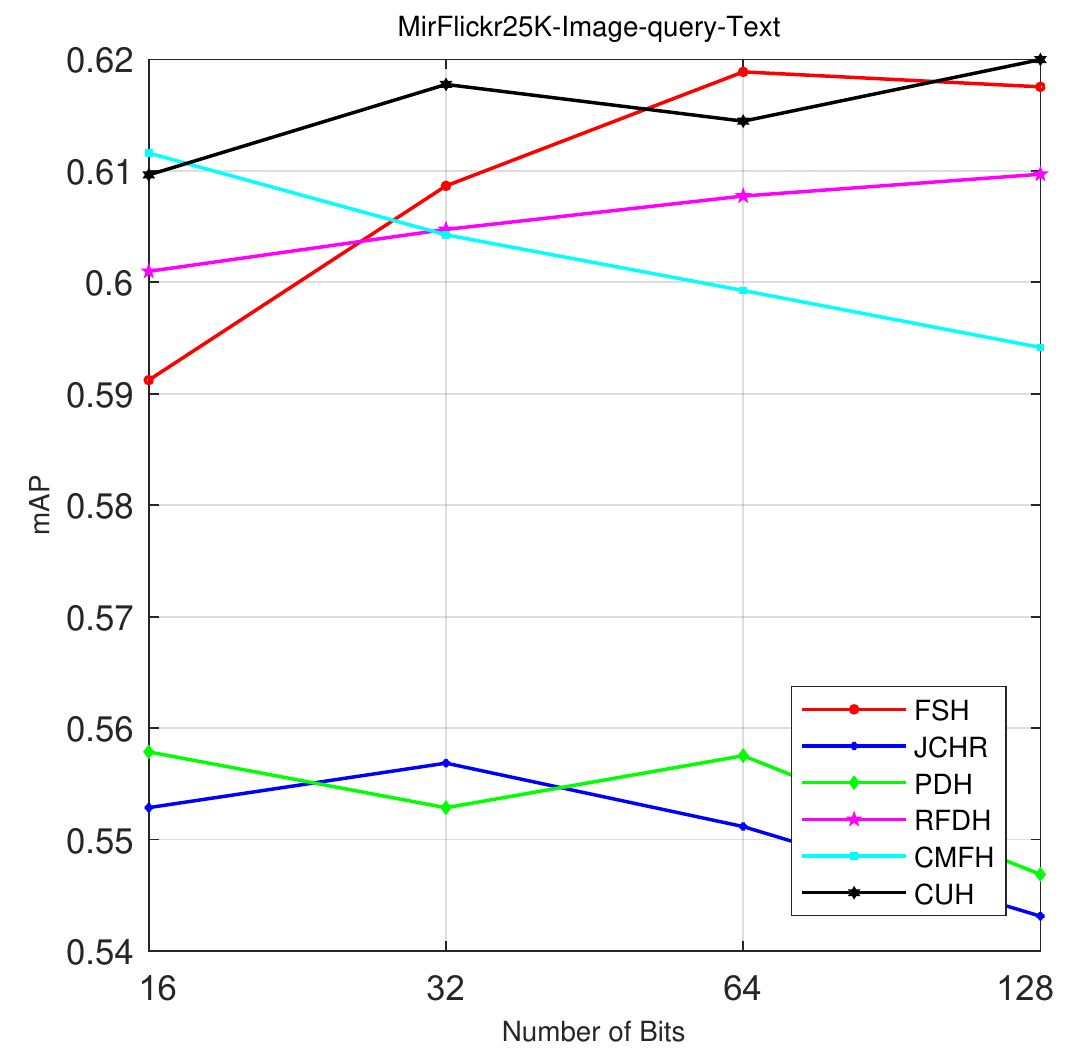}}
\hfill
\centering
\subfigure{
  \includegraphics[width=.2\textwidth]{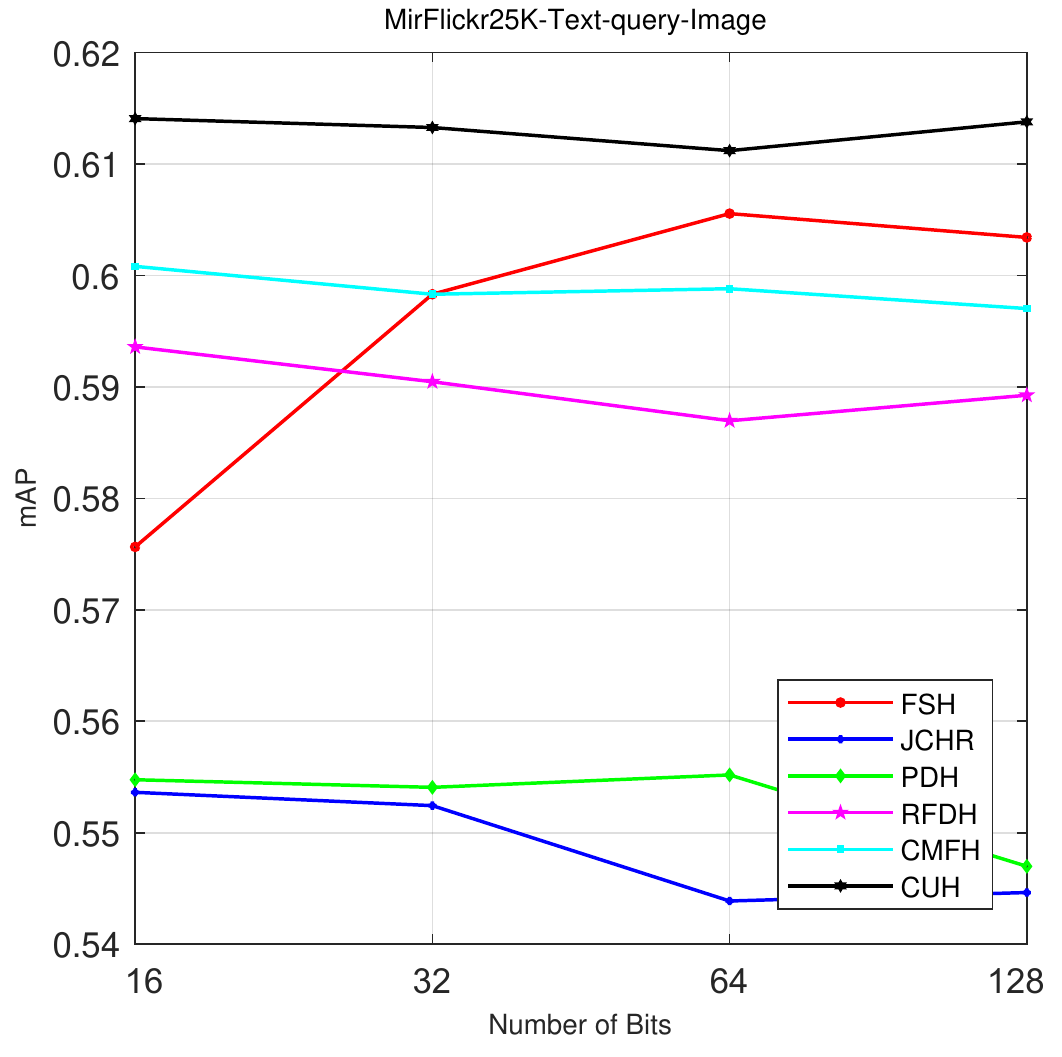}}
\caption{mAP values versus bits on MirFlickr25K.}
\label{figure6}
\end {figure}
\begin {figure} [t]
\centering
\subfigure{
  \includegraphics[width=.2\textwidth]{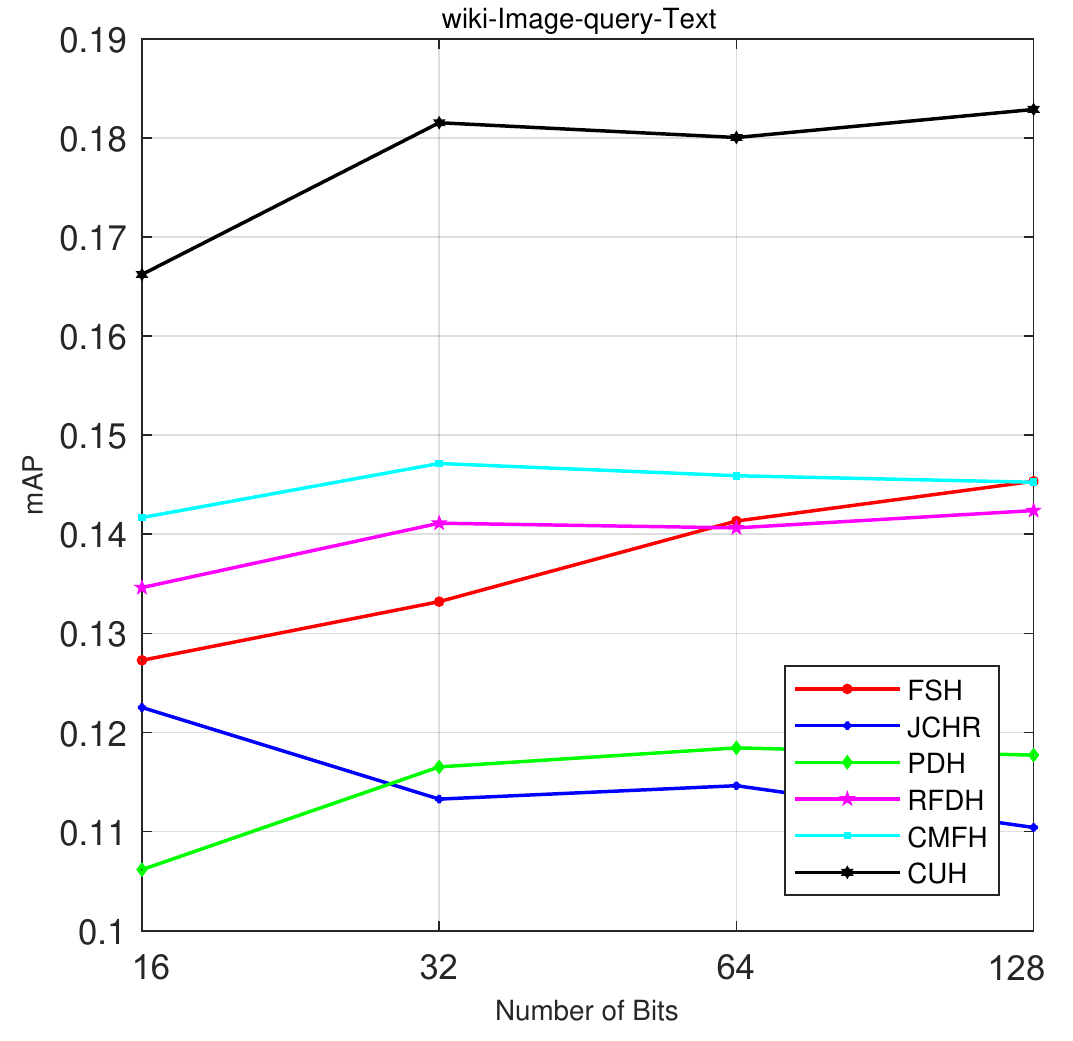}}
\hfill
\centering
\subfigure{
  \includegraphics[width=.2\textwidth]{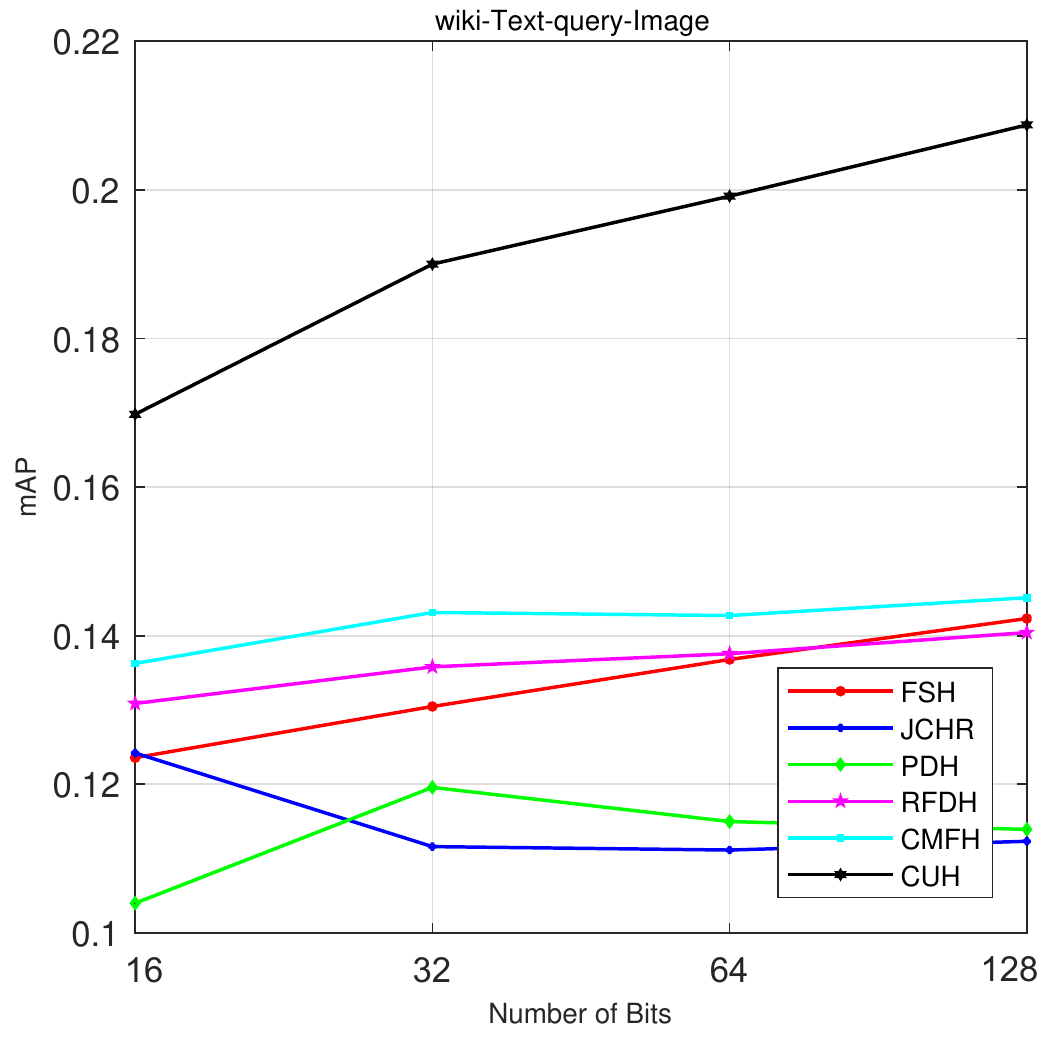}}
\caption{mAP values versus bits on Wiki.}
\label{figure7}
\end {figure}

The effectiveness and efficiency of the proposed CUH model are  conducted on three multimodal benchmark data sets: Wiki \cite{jour1}, MIRFlickr25K \cite{jour2}, NUS-WIDE \cite{jour3}. Specifically, some statistical characteristic of all data sets are depicted in the following.

Wiki \cite{jour1} consists of $2,866$ image-text pairs collected from Wikipedia¡¯s articles. It is grouped into $10$ semantic categories, where each image-text pair is belong to one of the $10$ semantic concepts in the categories. It makes a $128$-dimensional bag-of-visual-words vector constructed from the SIFT feature to represent every image, and a $10$-dimensional topics vector learned by a latent Dirichlet allocation (LDA) model to represent every text. In Wiki, a training set contains $2,173$ image-text pairs
randomly selected from the whole data set, and a query set contains $693$ image-text pairs with the remaining pairs. Besides, training set also is used as the database for retrieval evaluation. In the evaluation, we define the true semantic neighbors for a query through the associated labels.

MIRFlickr25K \cite{jour2} comprises $25,000$ images related to their tags from the Flickr website, in which all image-tag pairs are itemized into the $24$ semantic categories. In addition, one pair may have multiple labels from some of the $24$ semantic categories above. In the evaluation, we discard image-tag pairs that do not have tags or manually annotated labels and only select tags that appear at least $20$ times. By this pretreatment, we can employ a multimodal benchmark of $20,015$ image-tag pairs in experiment. It represents an image by a $150$-dimensional edge histogram vector, and a text by a $500$-dimensional vector extracted by PCA transforming the tag index vector in each pair. In MIRFlickr25K, the query set of $2000$ image-tag pairs are randomly taken from the whole data set, and the left pairs are used as the training set, which also serve as the database. In the evaluation, we define the true semantic neighbors for a query as those having at least one label with it at the same time.

NUS-WIDE \cite{jour3} contains $269,648$ images, which are downloaded from real-world website Flickr, it also collecting over $5,000$ tags from user. There are $81$ concepts fully labeled in the entire data set for performance evaluation. Following \cite{jour3}, we only keep the image-tag pairs belonged to one of the $10$ most frequent concepts, and the whole data set is pruned as a new data set comprising $186,577$ image-tag pairs. In the experiments, $500$-dimensional bag-of-visual-words feature vector is choose to represent image, and an index feature vector of the most common $1,000$ tags is choose to represent text. In NUS-WIDE, the query set contains $2,000$ image-tag pairs randomly taken from $186,577$ image-tag pairs, and the remaining $184,577$ pairs are treated as the database. Besides, we randomly selected $5000$ image-text pairs as the training set, which is used to learn the hash model. In the evaluation, we define the true semantic neighbors for a query as those having at least one label with it at the same time.
\subsection{Evaluation Criteria}
To perform a fair evaluation, three widely metrics mAP, topN-precision, and precision-recall are adopt in the evaluation of the retrieval performance for the proposed method and comparison methods. The definitions of these three metrics are as follows:
\begin{enumerate}[(1)]
\item mAP.
Given a query and a list of R retrieved documents, the value of its average precision (AP) is defined as
\begin{equation}\label{CUH17}
AP=\frac{1}{N} \sum_{k=1}^R P(k)\delta(r),
\end{equation}
where $N$ is the number of relevant documents in retrieved set, $P(k)$ denotes the precision of the top $k$ retrieved documents, and $\delta(r)=1$ if the $k$-th retrieved document is a true neighbor of the query, and otherwise $\delta(r)=0$. Then the APs of all queries are averaged to obtain the mAP. R is set to $1000$ in the following experiments.

\item topN-precision.
It expresses the variation of precision with respect to the number of retrieved instances.

\item precision-recall.
It conveys the precision at different recall level, which can be gotten by changing the Hamming radius of retrieval and evaluating the precision and recall at the same time.
\end{enumerate}
In general, the larger the values of three popular metrics are, the better the performance. Detailed description of the above evaluation criteria can be referred to \cite{proceeding19}.
\subsection{Baseline Methods}
The proposed CUH model is compared with the following five state-of-the-art unsupervised multimodal hashing methods: PDH \cite{proceeding10}, CMFH \cite{proceeding11}, RFDH \cite{jour3}, FSH \cite{proceeding9}, JCHR \cite{proceeding20}. The parameters in above methods are set according to the corresponding papers.
\subsection{Implementation Details}
Initialization. Following \cite{proceeding18}, we will use a stable method to initialize G. Hence, we initialize G as follows:
\begin{equation}\label{CUH18}
G=1\otimes Z_C,
\end{equation}
where $I_C \in \Re^{C\times C}$ is a identity matrix and $Z_C \in \Re^{C\times C}$ is a binary matrix by randomly sorting the rows of $I_C$, and $1 \in \Re^{\left \lfloor N \div C \right \rfloor \times 1}$ is a column vector with all elements being $1$. This method uses direct product of vector $1$ and matrix $Z_C$ to initialize $G$. If $N$ cannot be divisible by $C$, we need to extra select $r=N-C\times \left \lfloor N \div C \right \rfloor$ rows from $Z_C$ randomly to fill the indivisible part. This initialization method can make the mapping relationships between different labels of different categories nearly invariable, which can lead to a more stable initialization. In our experiment, for all datasets, we applied this new initialization on all methods.

Parameter setting. The CUH model is related to three model parameters: the quantization error hyper-parameter $\lambda$, the cluster-wise code-prototypes regularization hyper-parameter $\beta$ and the number of cluster centroid points hyper-parameter numCluster. For CUH, the quantization error hyper-parameter $\lambda$ is set to $10^{-1}$, the cluster-wise code-prototypes regularization hyper-parameter $\beta$ is set to $10^{-4}$, and  the number of cluster centroid points hyper-parameter numCluster is set to $40$ throughout the comparative study. We will study parameter sensitivity in Section 4.7 to validate that CUH can consistently outperform the state of the arts with a wide range of parameter configurations.
\subsection{Results and Discussions}
In Fig. \ref{figure5}, \ref{figure6}, and \ref{figure7}, the mAP evaluation results are exhibited on all three data sets, i.e. Wiki, MIRFlickr25K, and NUS-WIDE respectively. From these figures, for all cross-modal tasks (i.e. image-query-text and text-query-image), CUH achieves significantly better result than all comparison methods On Wiki and MIRFlickr25K. Besides, CUH also shows comparable performance with CMFH, outperforming other remaining comparison methods on NUS-WIDE. Superiority of CUH can be attributed to their capability to reduce the effect of information loss, adjust the weights adaptively, as well as avoid the large quantization error. The above observations show the effectiveness of the proposed CUH.

\begin {figure} [t]
\centering
\subfigure{
  \includegraphics[width=.2\textwidth]{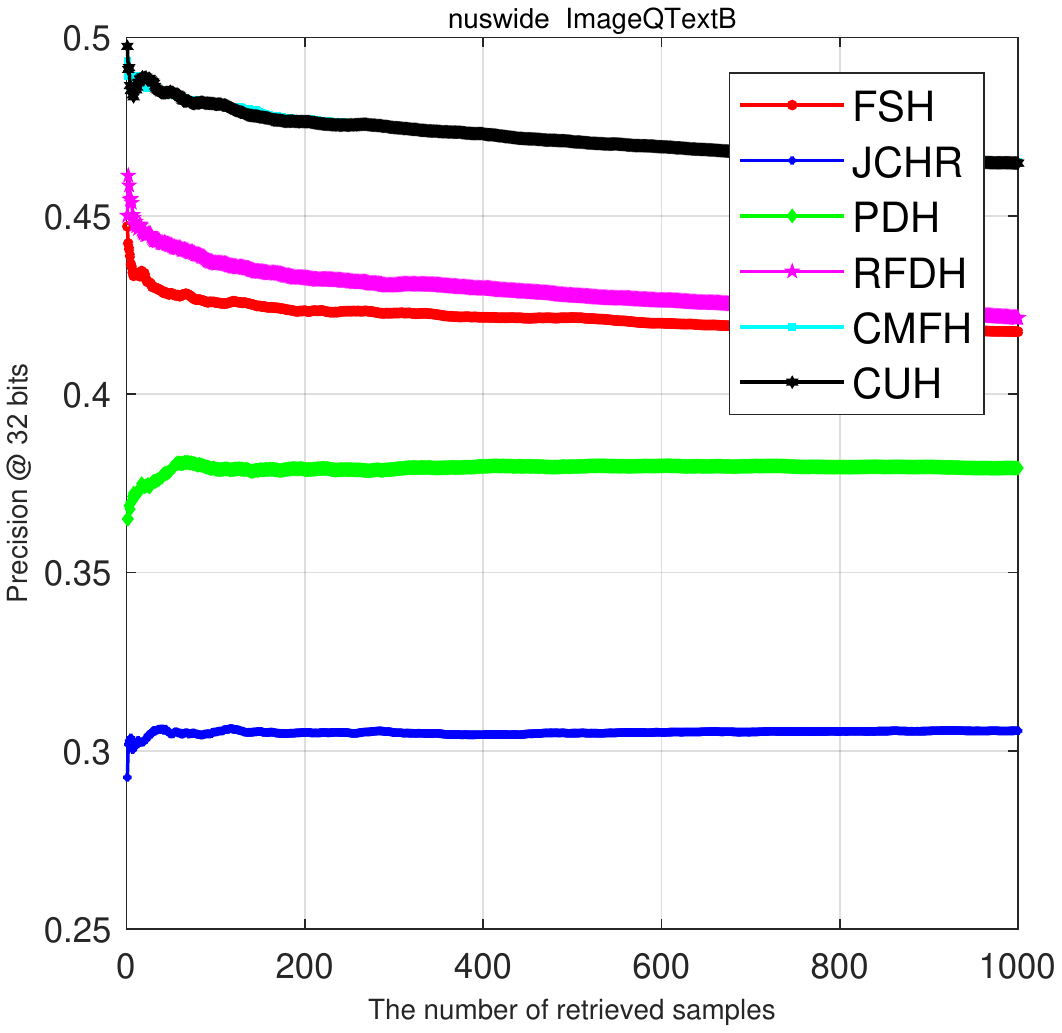}}
\hfill
\centering
\subfigure{
  \includegraphics[width=.2\textwidth]{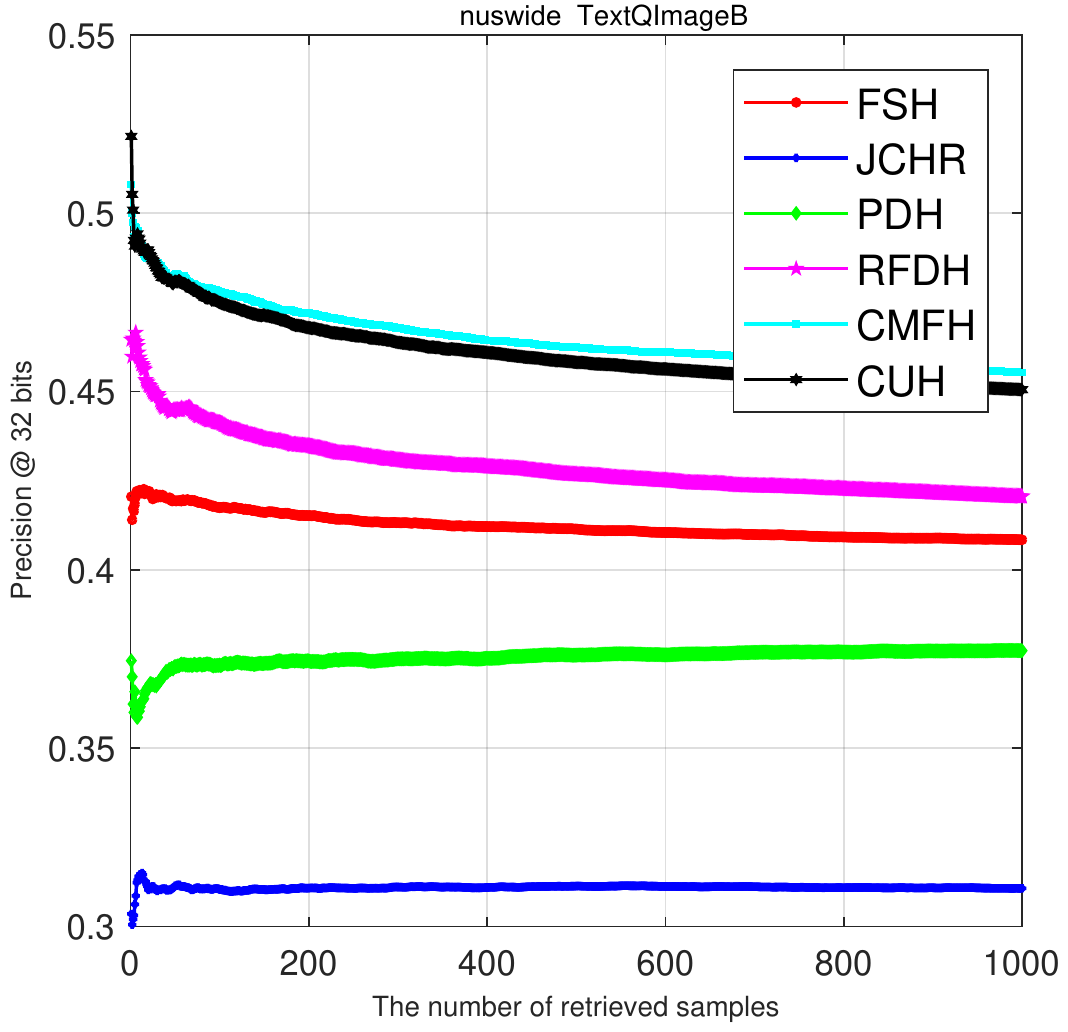}}
\caption{TopN-precision Curves @ $32$ bits on Nuswide.}
\label{figure11}
\end {figure}
\begin {figure} [t]
\centering
\subfigure{
  \includegraphics[width=.2\textwidth]{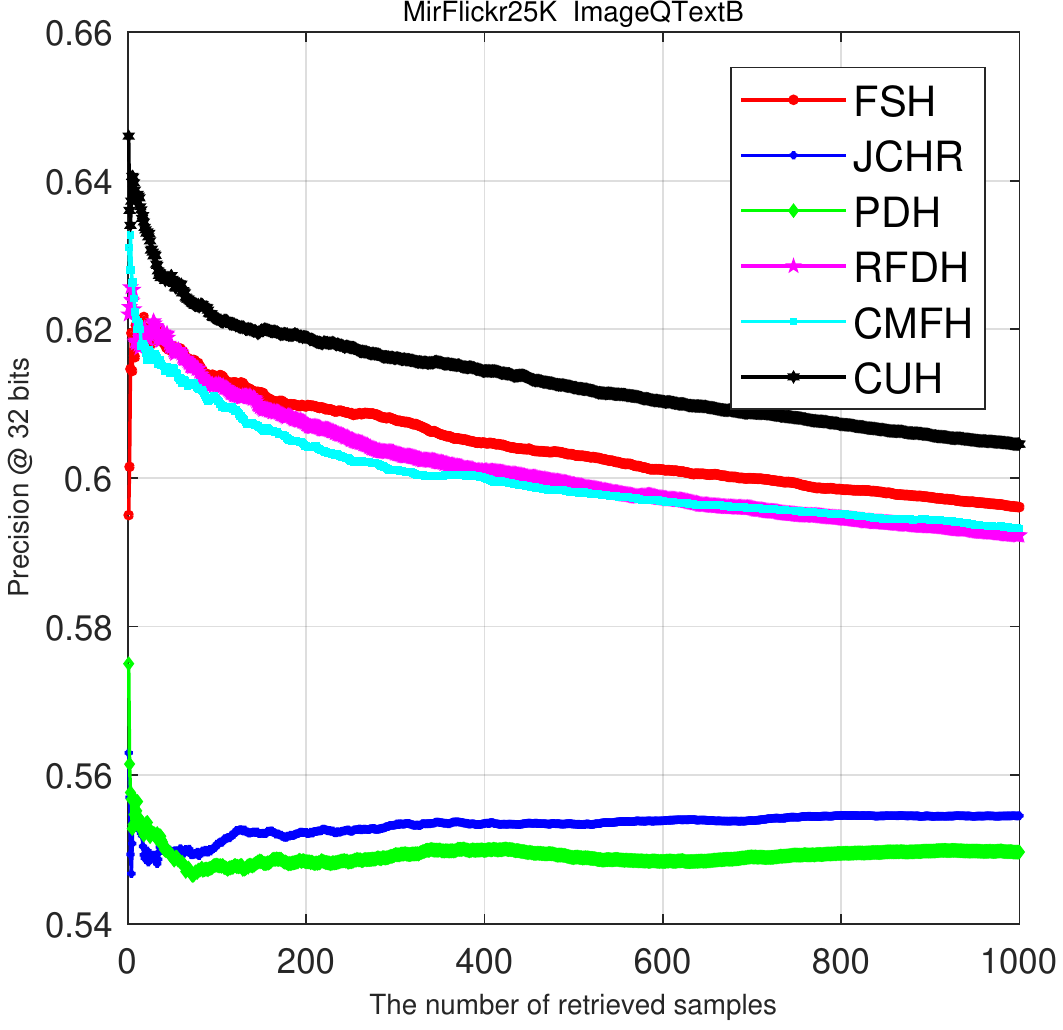}}
\hfill
\centering
\subfigure{
  \includegraphics[width=.2\textwidth]{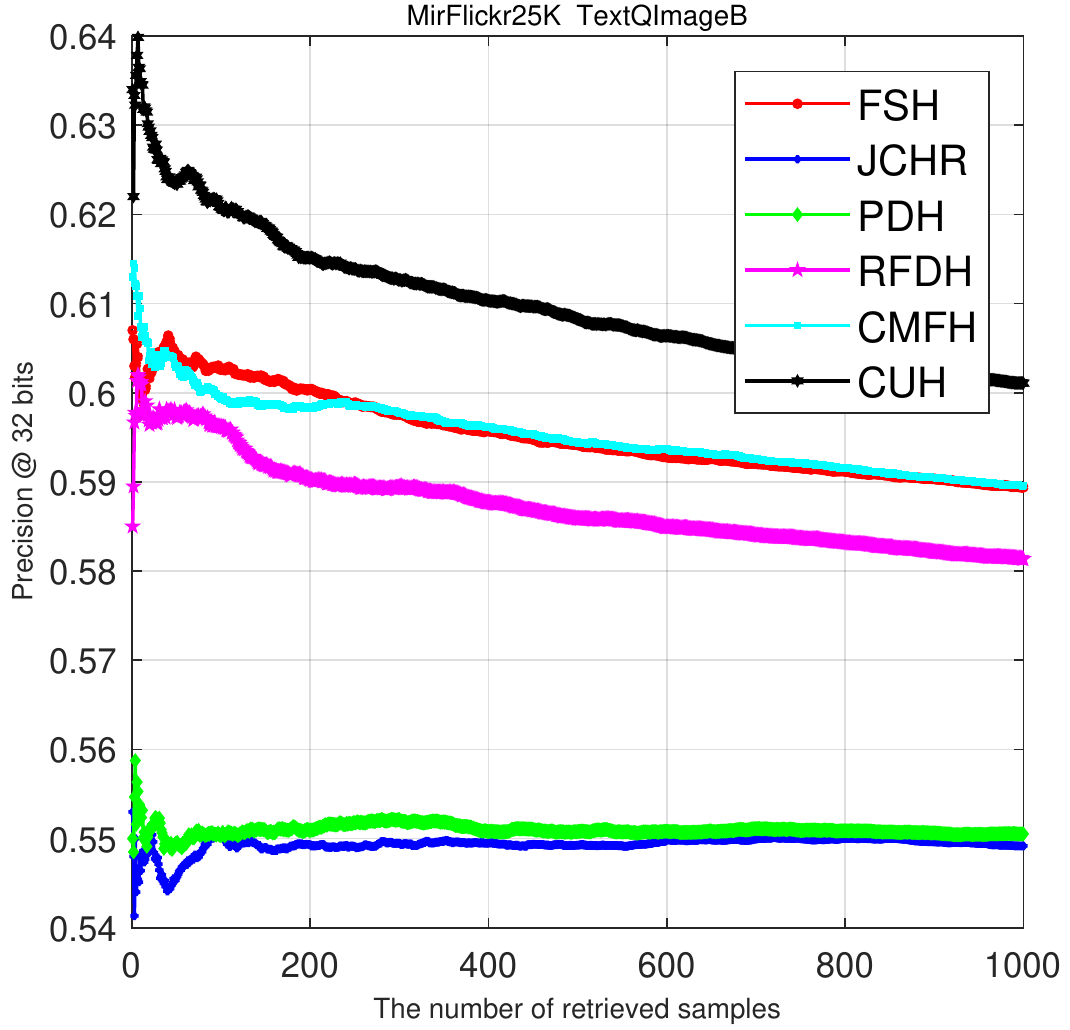}}
\caption{TopN-precision Curves @ $32$ bits on MirFlickr25K.}
\label{figure12}
\end {figure}
\begin {figure} [t]
\centering
\subfigure{
  \includegraphics[width=.2\textwidth]{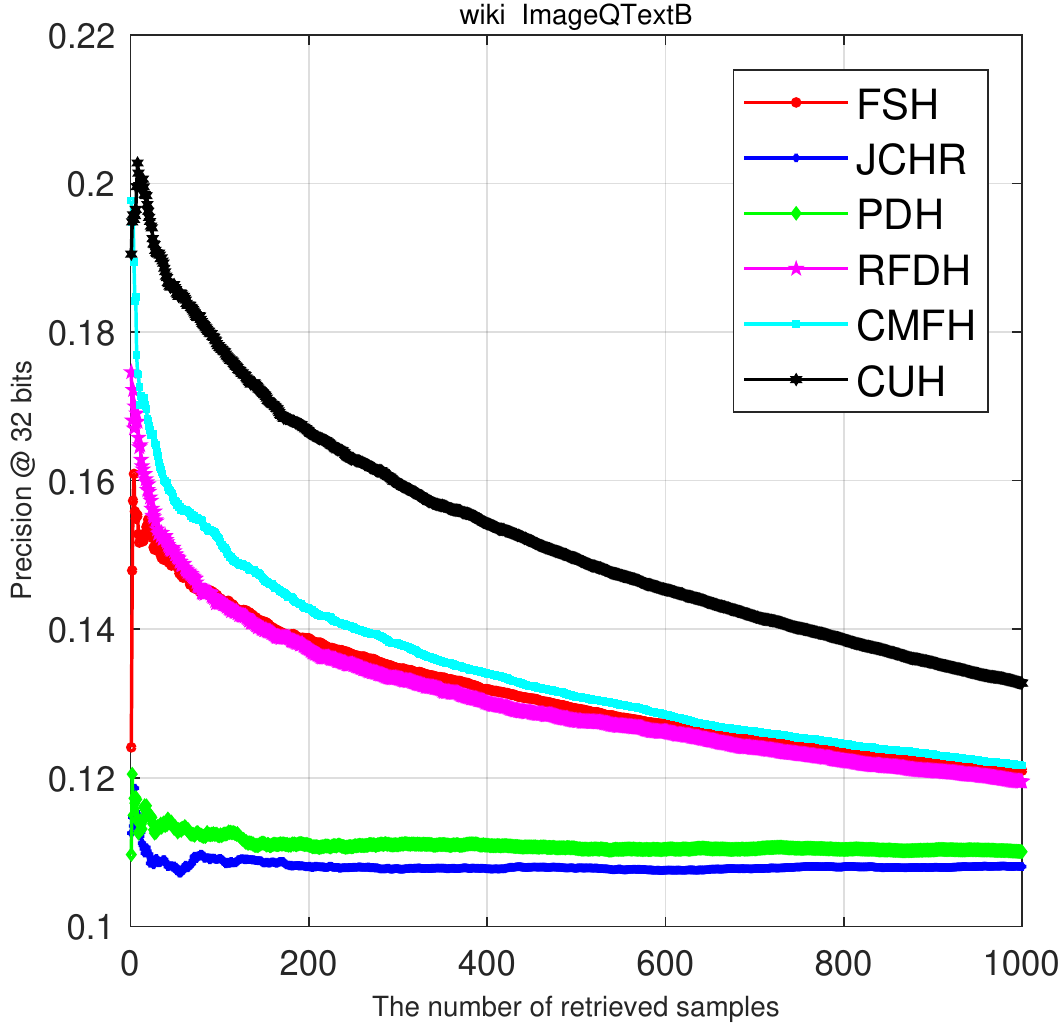}}
\hfill
\centering
\subfigure{
  \includegraphics[width=.2\textwidth]{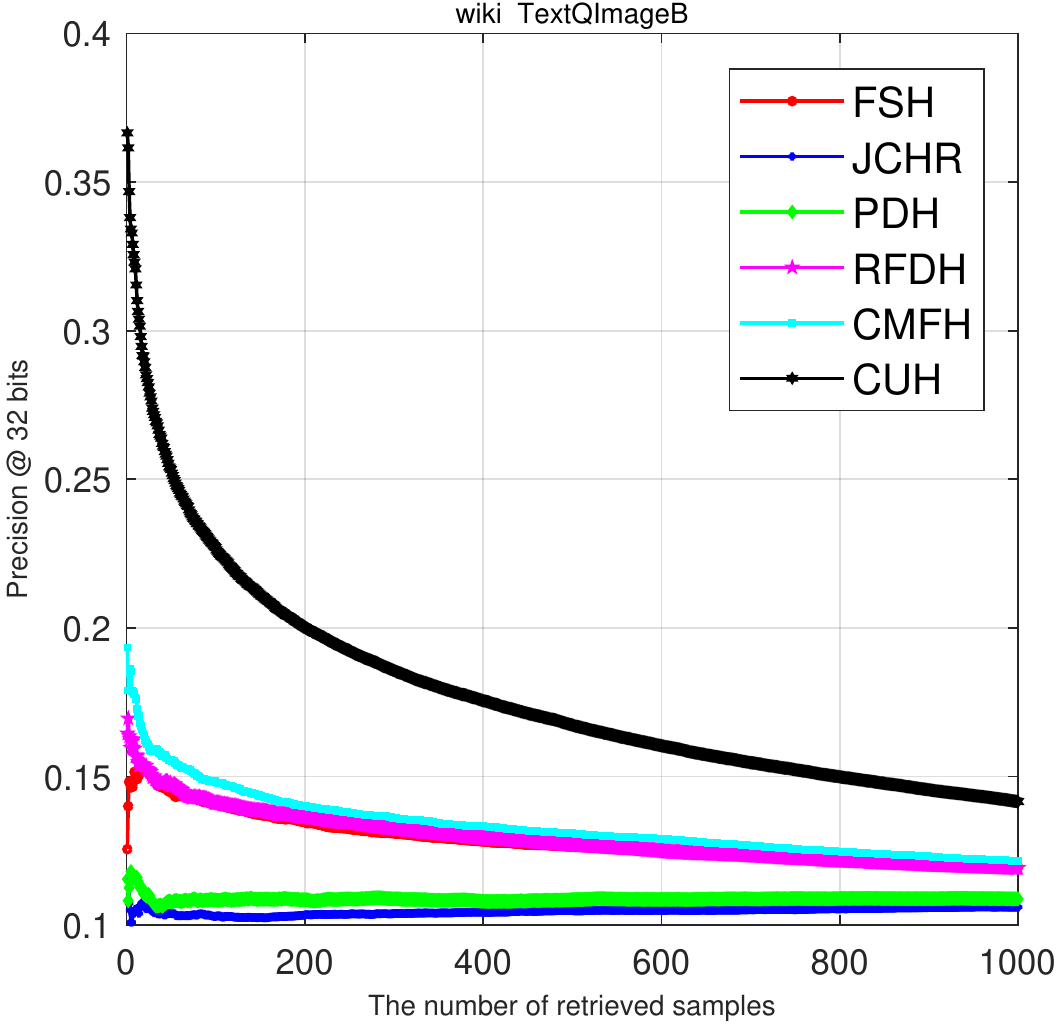}}
\caption{TopN-precision Curves @ $32$ bits on Wiki.}
\label{figure13}
\end {figure}
\begin {figure} [t]
\centering
\subfigure{
  \includegraphics[width=.2\textwidth]{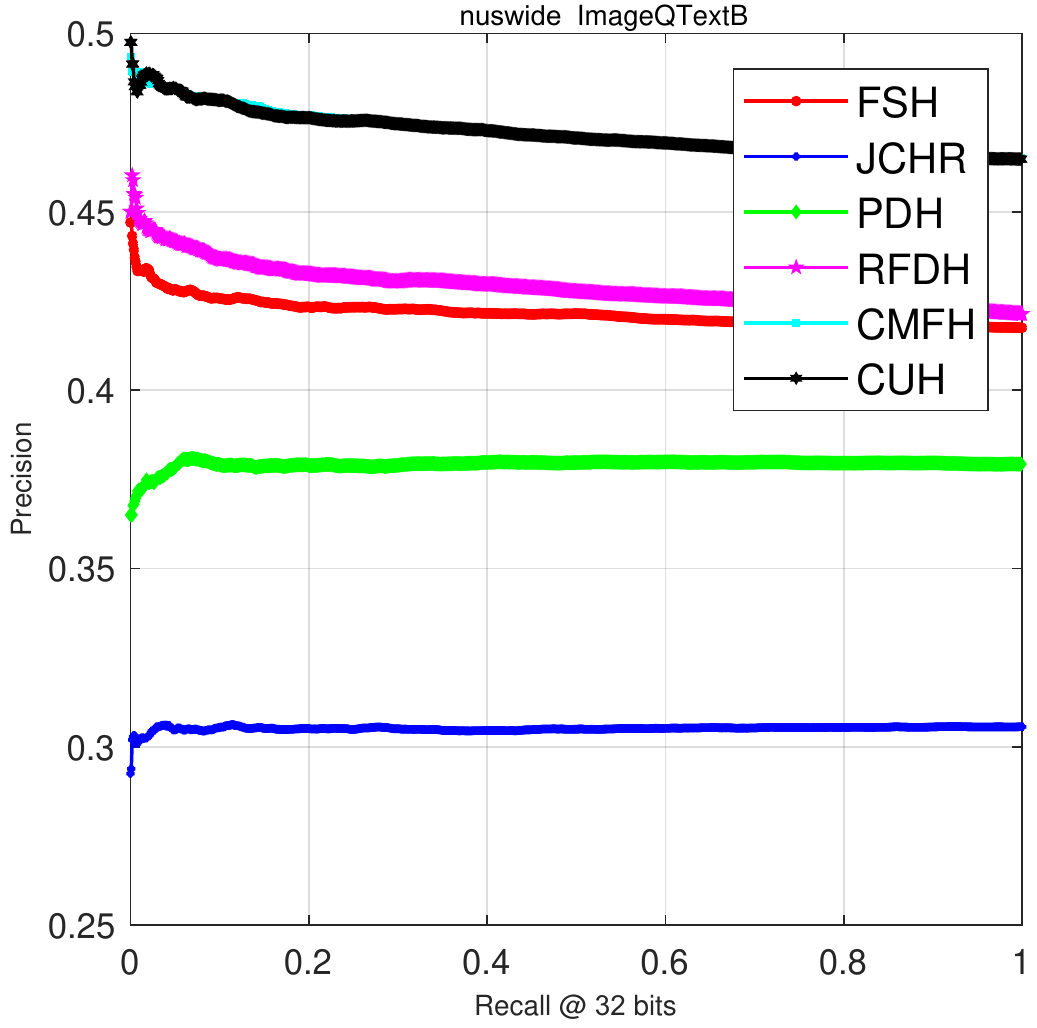}}
\hfill
\centering
\subfigure{
  \includegraphics[width=.2\textwidth]{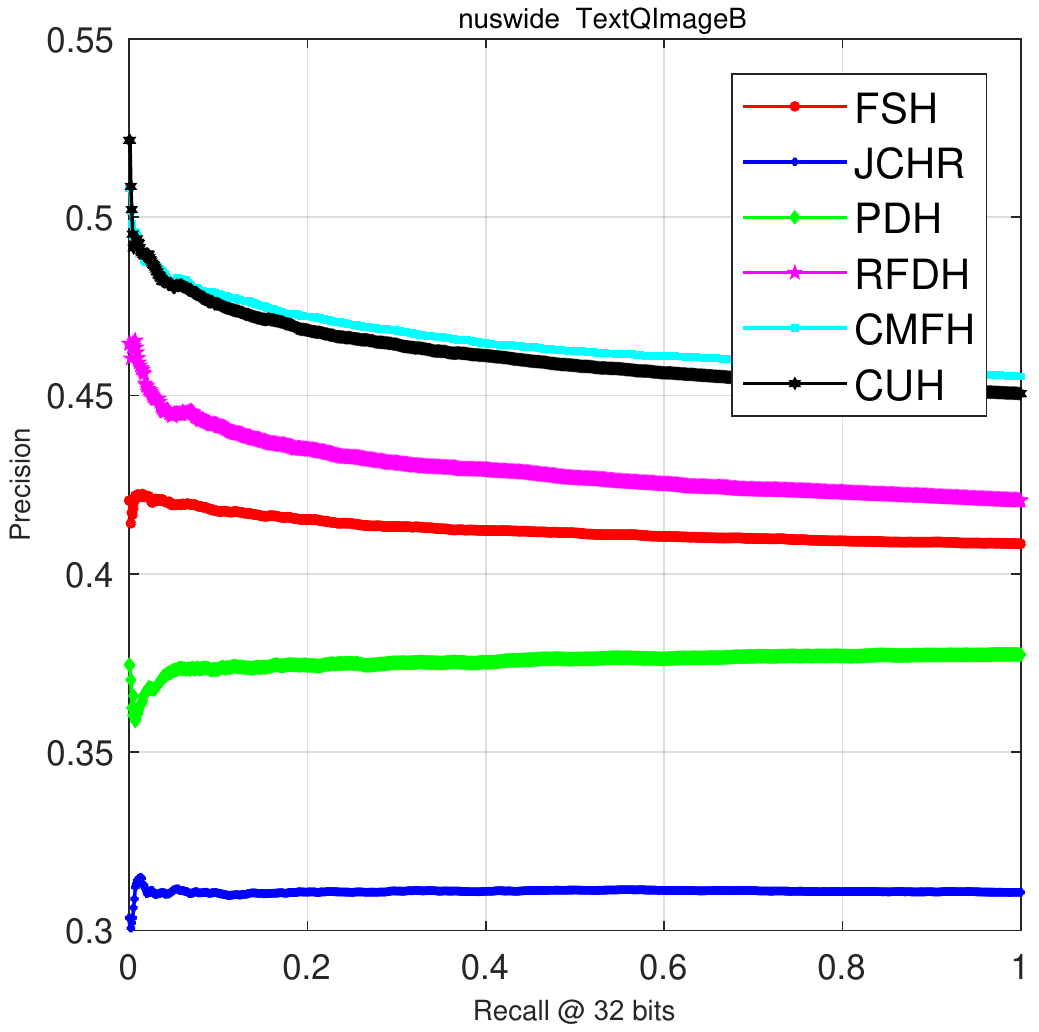}}
\caption{Precision-Recall Curves @ $32$ bits on Nuswide.}
\label{figure8}
\end {figure}
\begin {figure} [t]
\centering
\subfigure{
  \includegraphics[width=.2\textwidth]{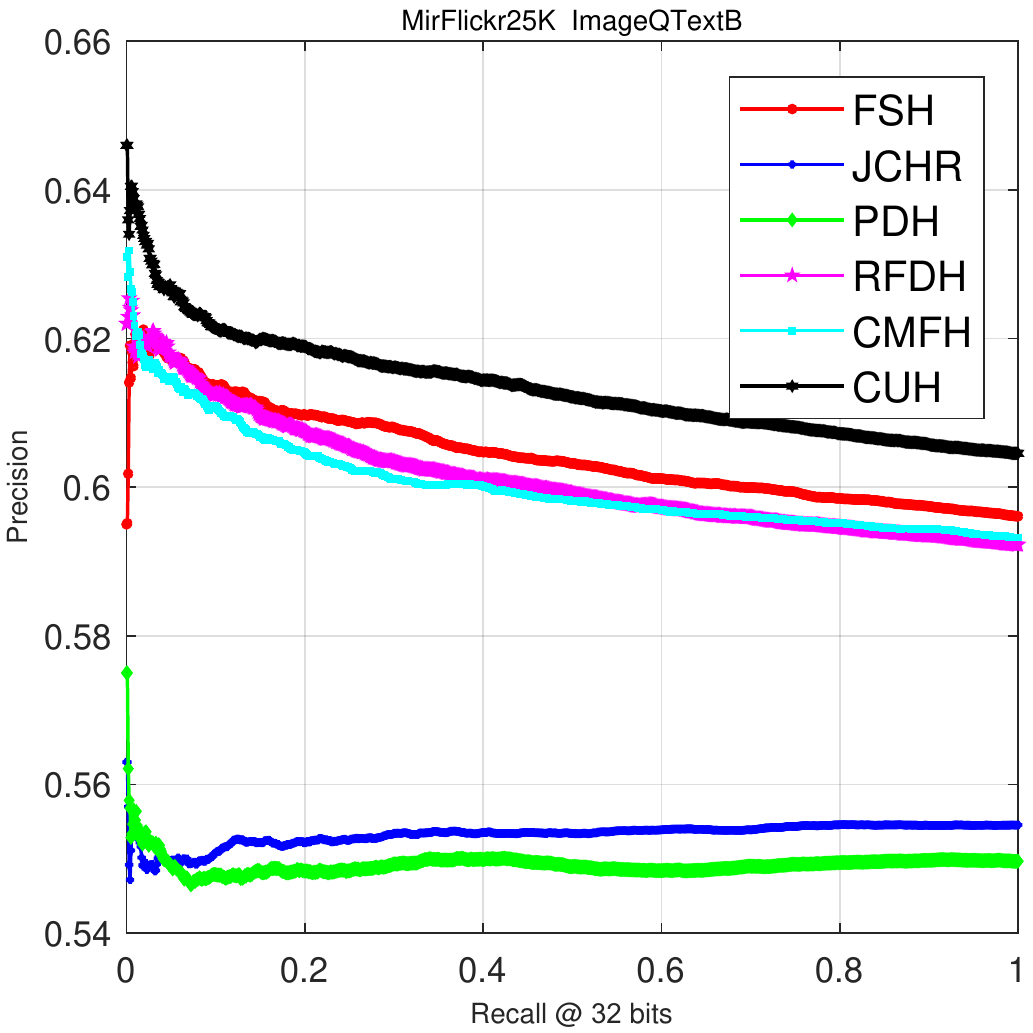}}
\hfill
\centering
\subfigure{
  \includegraphics[width=.2\textwidth]{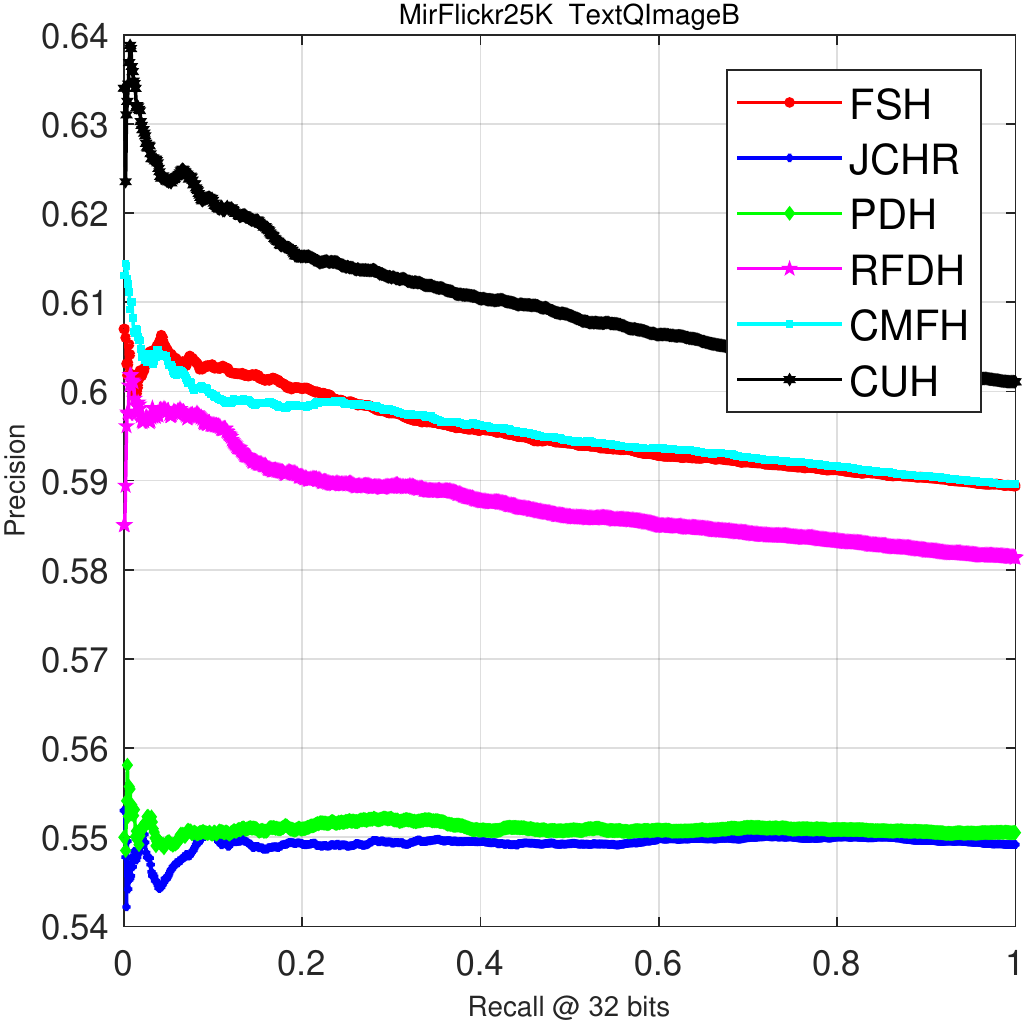}}
\caption{Precision-Recall Curves @ $32$ bits on MirFlickr25K.}
\label{figure9}
\end {figure}
\begin {figure} [t]
\centering
\subfigure{
  \includegraphics[width=.2\textwidth]{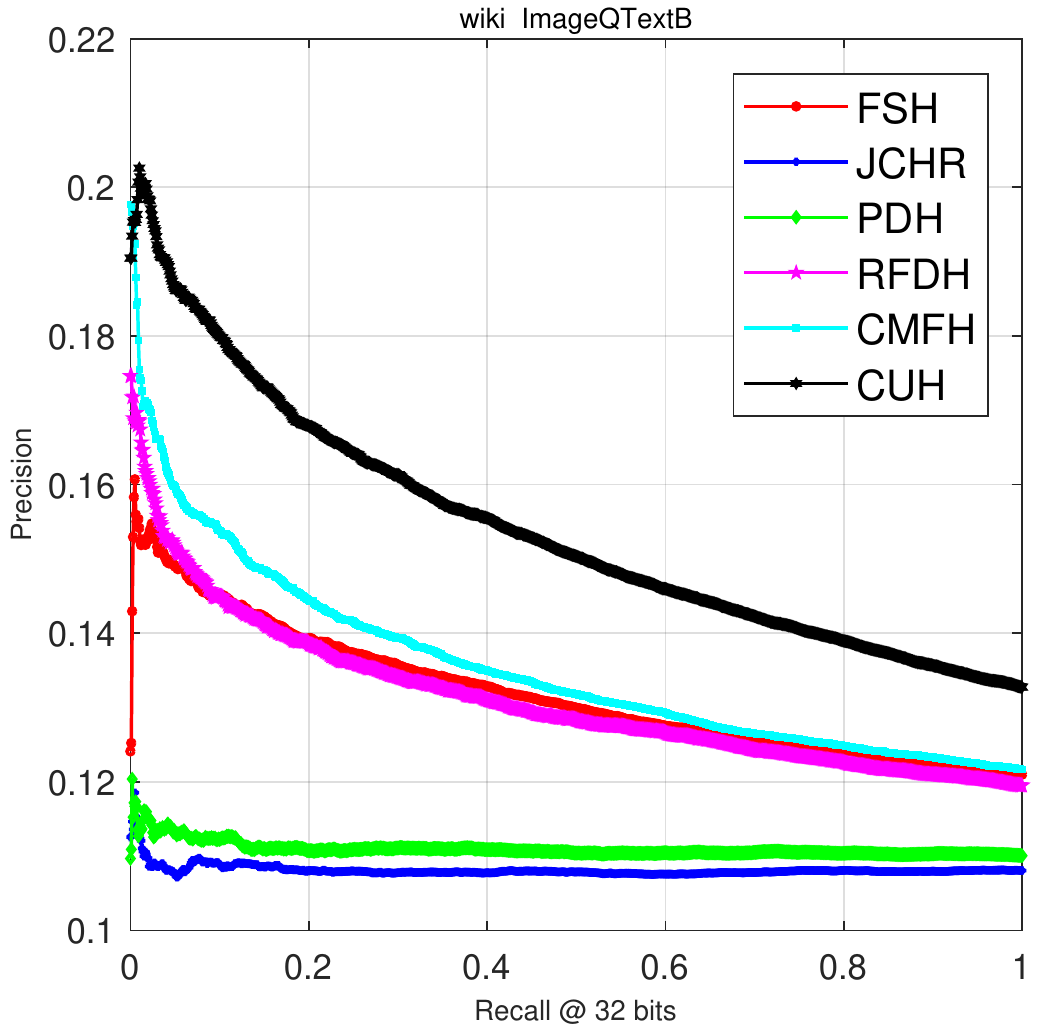}}
\hfill
\centering
\subfigure{
  \includegraphics[width=.2\textwidth]{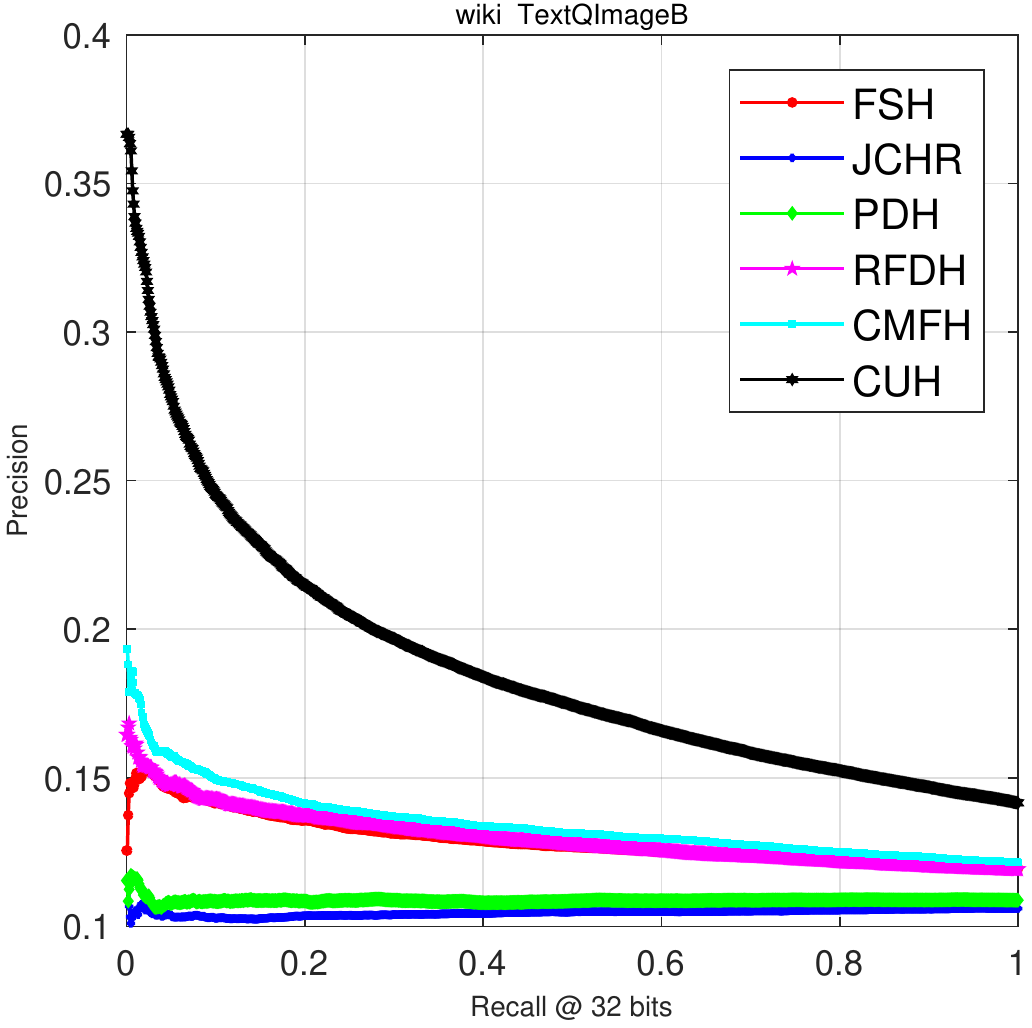}}
\caption{Precision-Recall Curves @ $32$ bits on Wiki.}
\label{figure10}
\end {figure}

The topN-precision curves with code length $32$ bits on all three data sets are demonstrated in Figs. \ref{figure11}, \ref{figure12}, and \ref{figure13} respectively. From the experimental results, the topN-precision results are in accordance with mAP evaluation values. CUH have better performance than others comparison methods about cross-modal hashing search tasks on Wiki and MIRFlickr25K. Further more, CUH demonstrates comparable performance with CMFH, outperforming other remaining comparison methods on NUS-WIDE. In retrieval system, we focus more on the front items in the retrieved list returned by search algorithm. Hence, CUH achieves better performance on all retrieval tasks in some sense.

From Figs. \ref{figure5}-\ref{figure13}, CUH usually demonstrates large margins on performance when compared with other methods about cross-modal hashing search tasks on Wiki and MIRFlickr25K. At the same time, CUH also exhibits comparable performance with CMFH, better than other remaining comparison methods on NUS-WIDE. We consider two possible reasons, explaining this phenomenon. First,
CUH utilizes least-absolute clustering residual in multi-view clustering for learning binary codes, which can be robust to data (i.e. image and text data) outliers and noises. Thus, CUH can achieve improvement on performance. Second, CUH keeps the inter-modal semantic coherence by multi-view clustering, which can
extract the  high-level hidden semantic features in the image and text. Therefore, CUH could find the common clustering indicators, that reflect the semantic properties more precise. On the consequences, under the guidance of the cluster-wise code-prototypes, CUH can achieve better performance on cross-modal retrieval tasks.

The precision-recall curves with code length of $32$ bits are also demonstrated in Fig. \ref{figure8}, \ref{figure9}, and \ref{figure10}. By calculation of the area under precision-recall curves, we can discover that CUH outperforms comparison methods about cross-modal hashing search tasks on Wiki and MIRFlickr25K. In addition, CUH has comparable performance with CMFH, better than other remaining comparison methods on NUS-WIDE.
\subsection{Convergence Analysis}
\begin {figure} [ht]
\centering
\includegraphics[width=.2\textwidth]{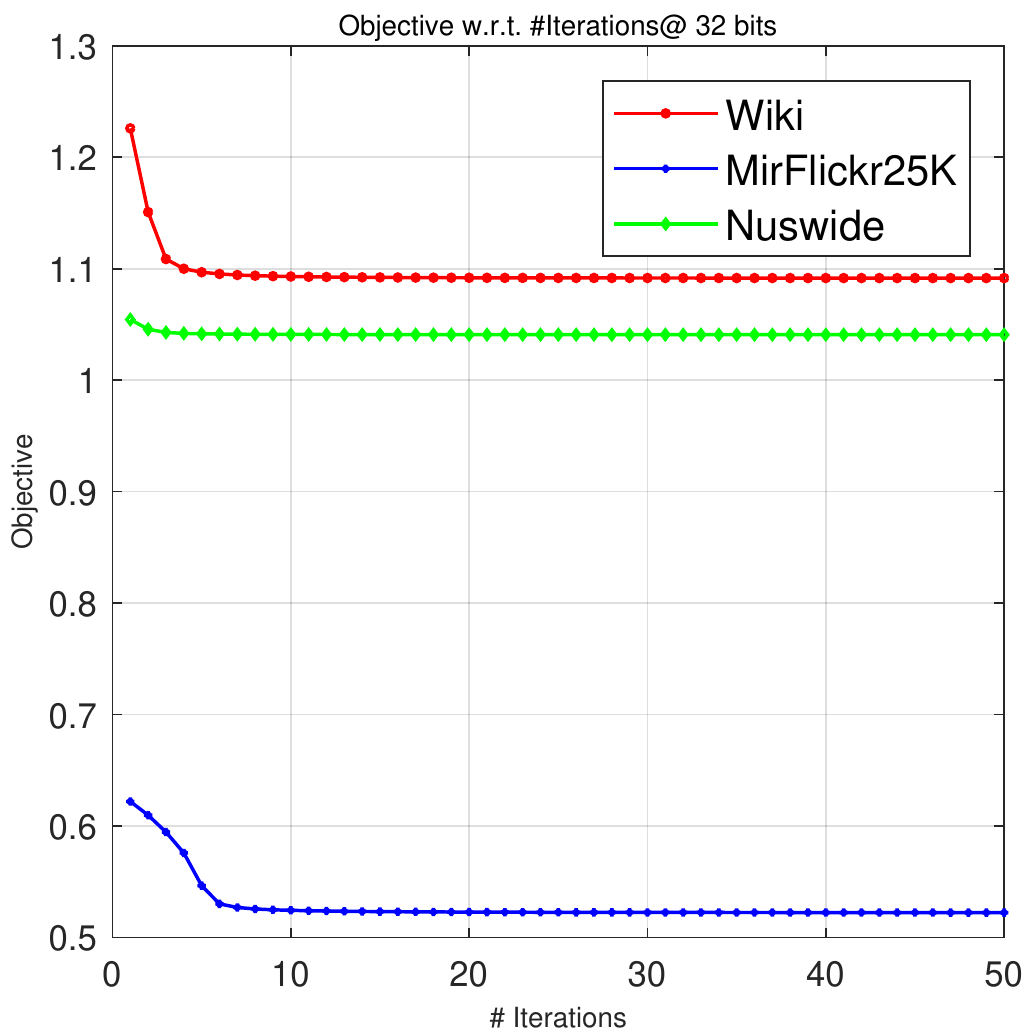}
\caption{Convergence Analysis.}
\label{figure1}
\end {figure}
\begin {figure} [t]
\centering
\subfigure{
  \includegraphics[width=.2\textwidth]{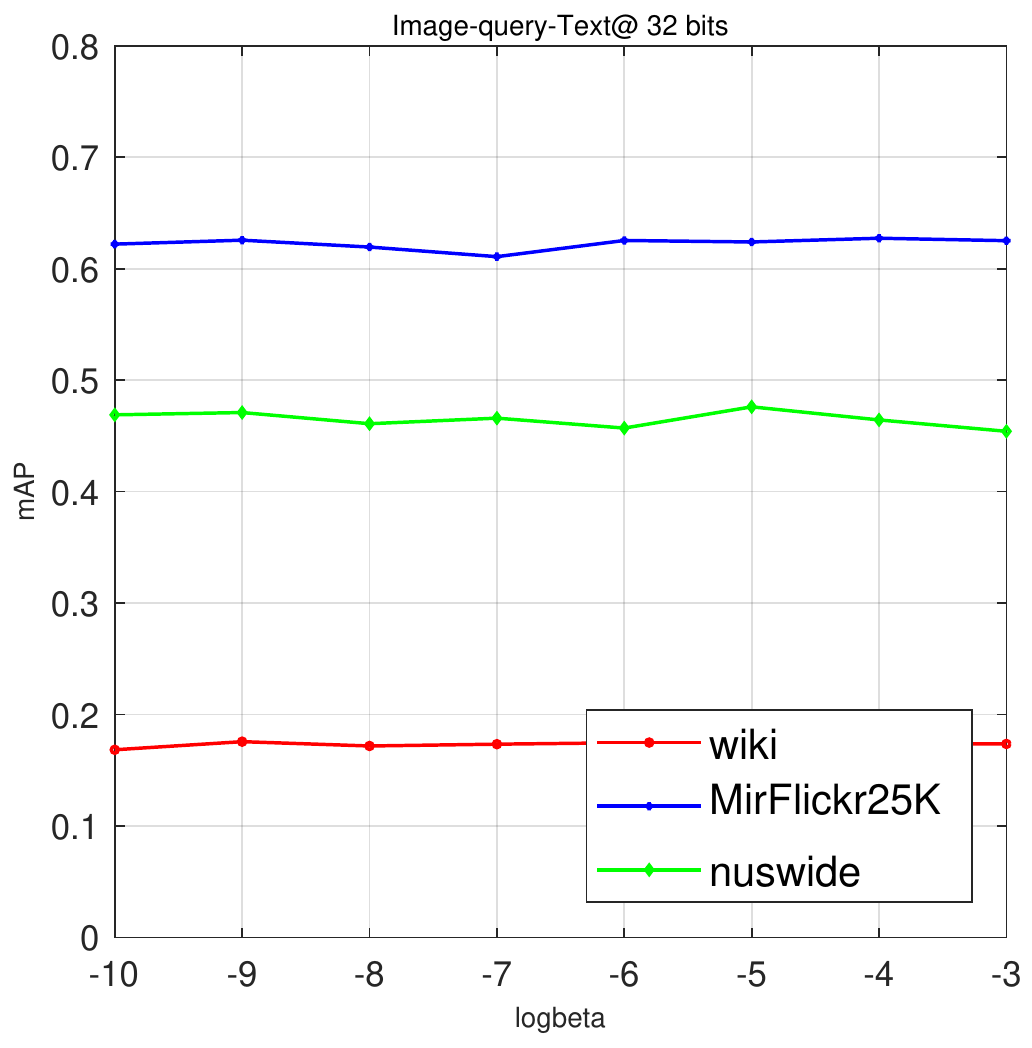}}
\hfill
\centering
\subfigure{
  \includegraphics[width=.2\textwidth]{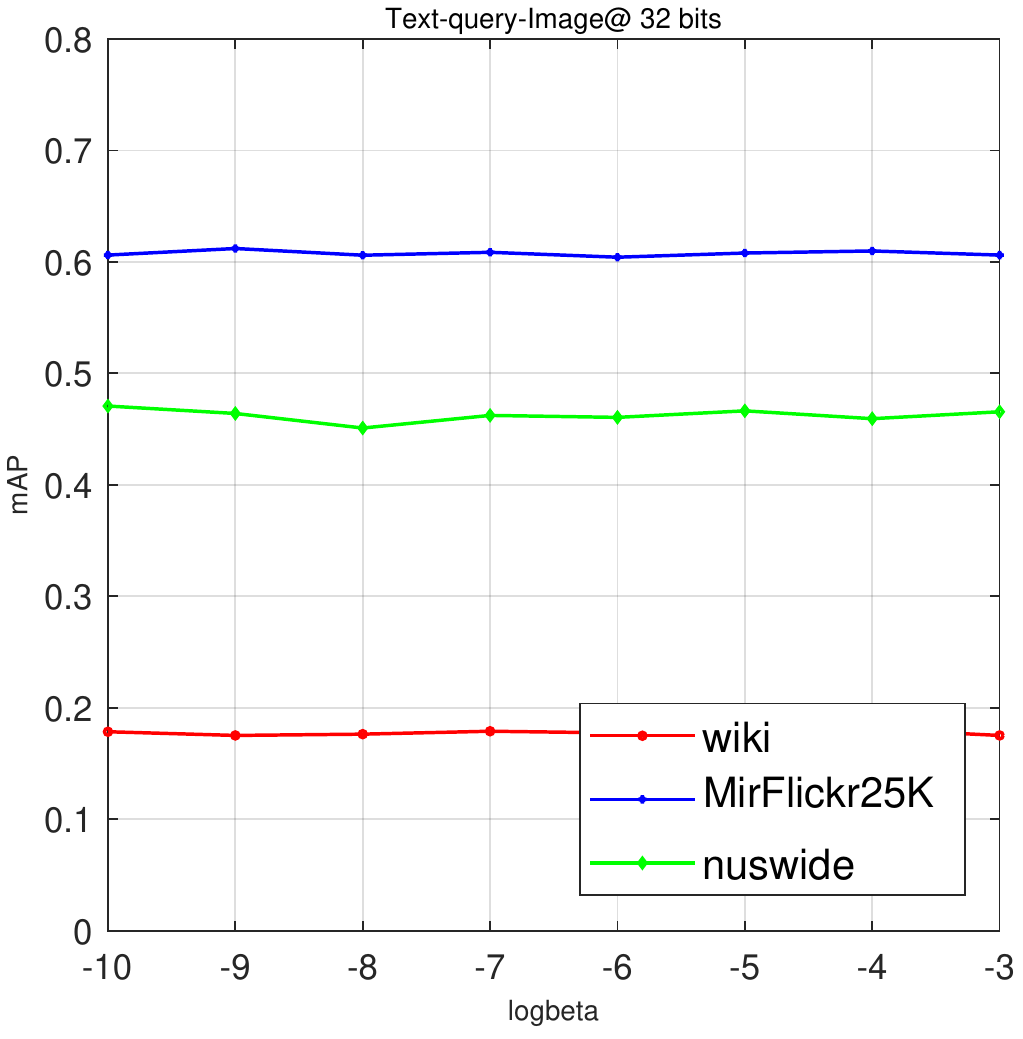}}
\caption{mAP values versus parameter $\beta $.}
\label{figure2}
\end {figure}
\begin {figure} [t]
\centering
\subfigure{
  \includegraphics[width=.2\textwidth]{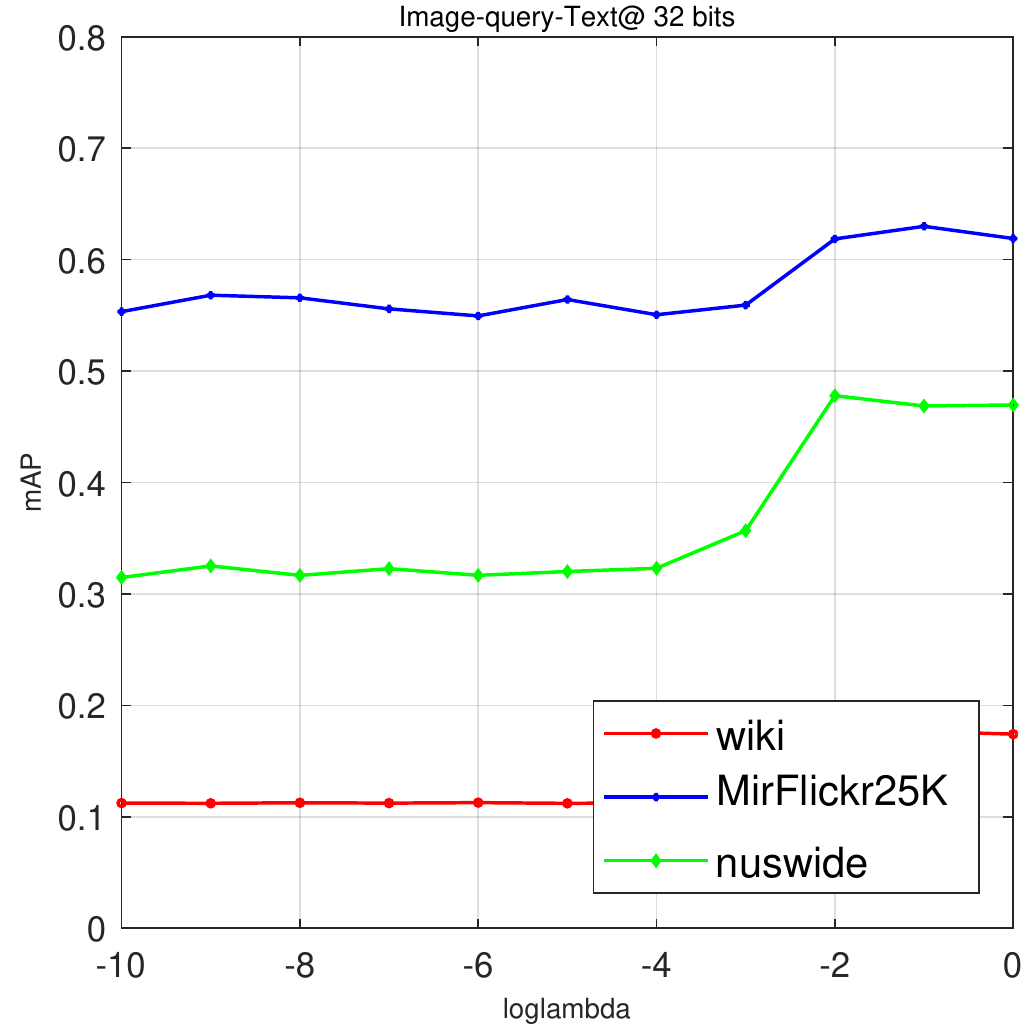}}
\hfill
\centering
\subfigure{
  \includegraphics[width=.2\textwidth]{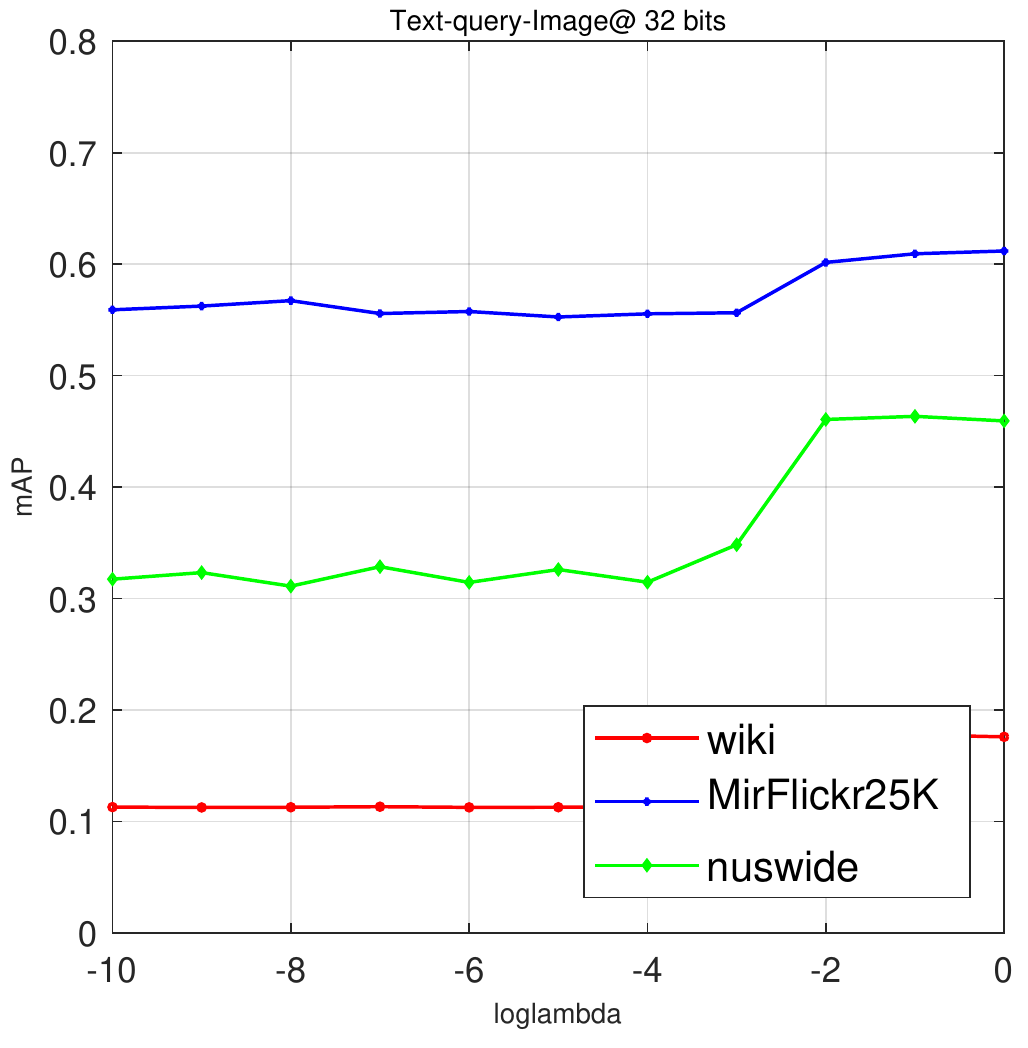}}
\caption{mAP values versus parameter $\lambda $.}
\label{figure3}
\end {figure}
\begin {figure} [t]
\centering
\subfigure{
  \includegraphics[width=.2\textwidth]{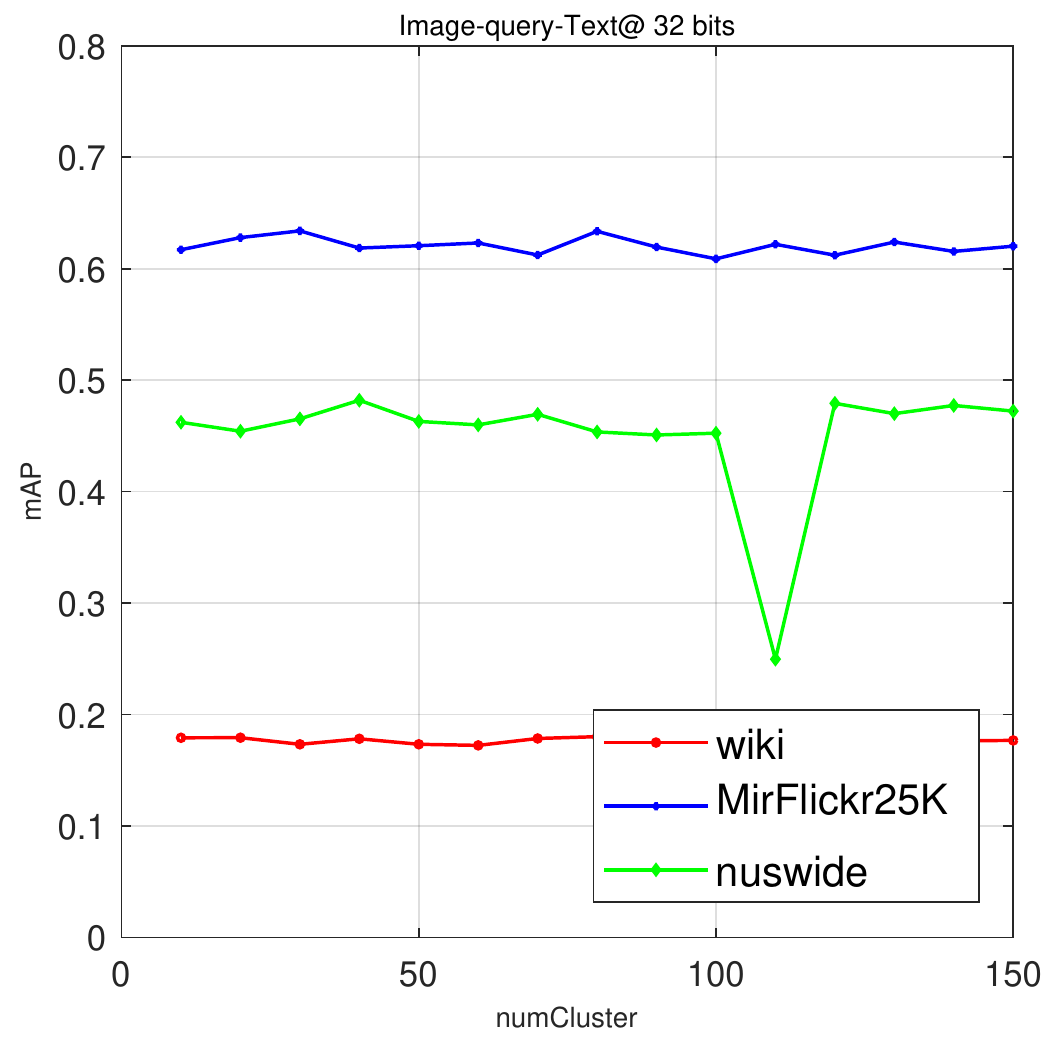}}
\hfill
\centering
\subfigure{
  \includegraphics[width=.2\textwidth]{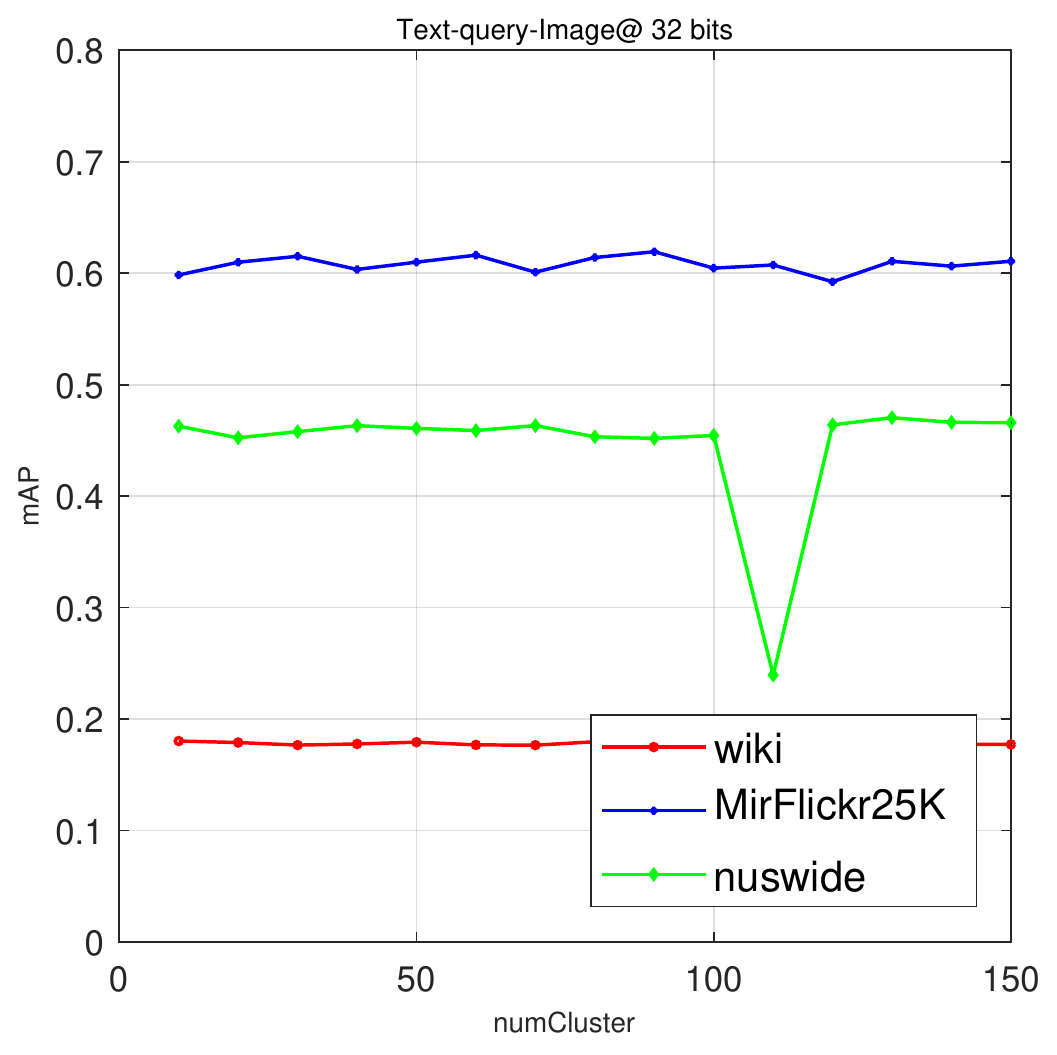}}
\caption{mAP values versus parameter $numCluster$.}
\label{figure4}
\end {figure}

Since CUH is solved in iterative steps, we empirically analyse its convergence property. Fig. \ref{figure1} demonstrates that the value of the objective (the value is averaged by the number of training data) can fall steadily with the number of iterations. From Fig. \ref{figure1}, we can realize the value of the objective will converge with $15$ iterations on all three datasets at $32$ bit. This result verifies the effectiveness of Algorithm \ref{alg:CUH}.
\subsection{Computational Complexity Analysis}
In this section, the train and test time about different cross-modal hashing methods measured in the study. The test time refer to the time that implements an out-of-sample binary code extension for all query and database instances. Our comparison is performed on a PC, which has configuration of 2.20GHz i7-8750H CPU and 16.0GB RAM. We evaluate the time cost of training and testing on the Wiki data set containing $2,173$ training pairs and on the NUS-WIDE data set consisting of $5,000$ training pairs.
\renewcommand{\arraystretch}{1.0}
\begin{table}[htb]
\tiny
  \centering
  \setlength{\belowcaptionskip}{5pt}
  \captionsetup{justification=centering}
  \caption{Training time (s) of different hashing methods on Wiki and NUS-WIDE at $32$ bits.}
  \label {Table.1}
  \resizebox{0.45\textwidth}{!}{
    \begin{tabular}{|c|c|c|}
    \hline
   \multirow{2}{*}{Method} &
    \multicolumn{1}{c|}{Wiki}&
    \multicolumn{1}{c|}{NUS-WIDE}\cr\cline{2-3}
    &Training time&Training time\cr
    \hline
    FSH &9.1582	&21.0636
\cr
   JCHR &4.4338 &9.6238
\cr
    PDH &25.8539	&137.0271
\cr
    RFDH &33.0569	&395.9939	
\cr
    CMFH &4.3356	&8.9293
\cr
    {\bf CUH} &9.8471	&20.8080
\cr
    \hline
    \end{tabular}}
\end{table}
\begin{table}[htb]
\tiny
  \centering
  \setlength{\belowcaptionskip}{5pt}
  \captionsetup{justification=centering}
  \caption{Testing time (s) of different hashing methods on Wiki and NUS-WIDE at $32$ bits.}
  \label {Table.2}
  \resizebox{0.45\textwidth}{!}{
    \begin{tabular}{|c|c|c|c|}
    \hline
    \multirow{2}{*}{Tasks}& \multirow{2}{*}{Method} &
    \multicolumn{1}{c|}{Wiki} &\multicolumn{1}{c|}{NUS-WIDE}\cr\cline{3-4}
    &&Testing time&Testing time\cr
    \hline
    \multirow{6}{*}{I$\rightarrow$T}
    &FSH &0.0506	&9.1641
\cr
   &JCHR &0.0501	&9.2376	
\cr
    &PDH &0.0502	&9.1492	
\cr
    &RFDH &0.0508	&9.1982
\cr
    &CMFH &0.0495	&9.3734
\cr
    &{\bf CUH} &0.0497	&9.6722
\cr
\hline
    \multirow{6}{*}{T$\rightarrow$I}
    &FSH &0.0529	&9.1838
\cr
   &JCHR &0.0506	&9.1974
\cr
    &PDH &0.0501	&9.1726	
\cr
    &RFDH &0.0526	&9.2026	
\cr
    &CMFH &0.0510	&9.3257	
\cr
    &{\bf CUH} &0.0508	&9.2291	
\cr
    \hline
    \end{tabular}}
\end{table}
The comparison of training and testing time complexity at $32$ bits is shown in Table. \ref{Table.1} and Table. \ref{Table.2}. As shown in Table. \ref{Table.1}, RFDH needs most time for learning the model, since it is a two-step learning scheme despite its good performance, in which it first learns binary codes, then trains hash functions. On the other hand, our CUH can learn the hash codes and hash functions in an acceptable speed, which also has better retrieval performance than existing cross-modal hashing methods. Besides, the testing time of comparison in Table. \ref{Table.2} is nearly identical for the compared cross-modal hashing model.
\subsection{Parameter Sensitivity Analysis}
In this section, we conduct the parameter sensitivity to verify that
the proposed CUH can achieve stable and superior performance under a large range of parameter values. We test the performance effects about the algorithm in different settings on all datasets. Here, we untilize mAP performance at $32$ bits for reporting the variation of performance with respect to parameter values. Our CUH has three hyper-parameters, which include the quantization error hyper-parameter $\lambda$, the cluster-wise code-prototypes regularization hyper-parameter $\beta$ and the number of cluster centroid points hyper-parameter numCluster.

The parameter $\lambda$ balances the reconstruction quantization error and clustering error in the CUH model. It can be observed from Fig. \ref{figure3} that the performance of CUH goes down slightly when $\lambda$ increasing. We find CUH can achieve best performance around $\lambda =10^{-1}$ on all three datasets. Fortunately, when we select $\lambda$ form the range $\left [ 10^{-2}, 1 \right ]$, the robust performance of the proposed CUH can be guaranteed.

The parameter $\beta$ is a hyper-parameter, which balances the cluster-wise code-prototypes regularization and clustering error in the CUH model. From Fig. \ref{figure2}, we can see that Wiki, MirFlickr25K and NUS-WIDE achieve the best around $\beta =10^{-4}$. Besides, we can observe that CUH achieves stable and superior performance under a large range of $\beta$.

The parameter numCluster is a hyper-parameter, which controls the number of cluster centroid points in the CUH model. From Fig. \ref{figure3}, we can see that Wiki, MirFlickr25K and NUS-WIDE achieve the best around $numCluster =40$. Besides, we can observe that CUH achieves stable and superior performance under a large range of $numCluster$.
\section{Conclusions}
In this paper, we have formally found a novel way out of cross-modal similarity retrieval task through the proposed cluster-wise unsupervised hashing (CUH) in the unsupervised case. It integrates multi-view clustering and learning of hash codes under the help of cluster-wise code-prototypes, i.e. cluster centroid points in multi-view clustering into an unified binary optimization framework, which generates better compact binary codes that sufficiently contain enough both inter-modal semantic coherence and intra-modal similarity. The binary codes across modalities are learnt under the guidance cluster-wise code-prototypes in its own latent semantic space, which is use for the purpose of key to the efficacy for the proposed CUH method. The reasonableness and effectiveness of CUH is well demonstrated by comprehensive experiments on diverse benchmark datasets. In the future, developing more non-linear mapping models such as boosting or a deep neural network seems an interesting work.
\section{References}
\bibliography{mybibfile}

\end{document}